\documentclass{IEEEtran}  % Comment this line out
\IEEEoverridecommandlockouts                              % This command is only
\usepackage[utf8]{inputenc}
\usepackage[T1]{fontenc}
\usepackage{graphicx}
\usepackage[namelimits]{amsmath} %数学公式
\usepackage{amssymb}             %数学公式
\usepackage{amsfonts}            %数学字体
\usepackage{mathrsfs}            %数学花体
\usepackage{subfigure} % 子图包
\usepackage{booktabs}
\usepackage{cite}
\usepackage{url}
\usepackage{dsfont}
\usepackage{algorithm}
\usepackage{algpseudocode}
\usepackage{algorithmicx}
\usepackage{amsmath}
\usepackage[switch]{lineno}
\usepackage{bbding}

\usepackage{hyperref}

\usepackage{multirow}
\usepackage{textcomp,booktabs}
\usepackage[usenames,dvipsnames]{color}
\usepackage{colortbl}
\usepackage[dvipsnames]{xcolor}
\definecolor{mgray}{gray}{.9}

\hypersetup{
    colorlinks=true,
    linkcolor=blue,
    filecolor=magenta,      
    urlcolor=cyan
}

\usepackage{soul,framed}

\usepackage[colorinlistoftodos,bordercolor=blue,backgroundcolor=blue!20,linecolor=blue,textsize=scriptsize]{todonotes}
\title{Hybrid-Prediction Integrated Planning for \\ Autonomous Driving}

\author{Haochen Liu, Zhiyu Huang, Wenhui Huang, Haohan Yang,\\
Xiaoyu Mo, and Chen Lv$^{*}$,~\IEEEmembership{Senior Member, IEEE}
\thanks{H. Liu, Z. Huang, W. Huang, H. Yang, X. Mo, and C. Lv are with the School of Mechanical and Aerospace Engineering, Nanyang Technological University, 639798, Singapore. (E-mails: {\tt \{haochen002, zhiyu001, wenhui001\}@e.ntu.edu.sg, \{haohan.yang, xiaoyu.mo, lyuchen\}@ntu.edu.sg})}
\thanks{$^{*}$Corresponding author: C. Lv}
}

\begin{document}
\maketitle
\thispagestyle{empty}
\pagestyle{empty}

%%%%%%%%%%%%%%%%%%%%%%%%%%%%%%%%%%%%%%%%%%%%%%%%%%%%%%%%%%%%%%%%%%%%%%%%%%%%%%%%
\begin{abstract}
Autonomous driving systems require the ability to fully understand and predict the surrounding environment to make informed decisions in complex scenarios. Recent advancements in learning-based systems have highlighted the importance of integrating prediction and planning modules. However, this integration has brought forth three major challenges: inherent trade-offs by sole prediction, consistency between prediction patterns, and social coherence in prediction and planning. To address these challenges, we introduce a hybrid-prediction integrated planning (HPP) system, which possesses three novelly designed modules. First, we introduce marginal-conditioned occupancy prediction to align joint occupancy with agent-wise perceptions. Our proposed MS-OccFormer module achieves multi-stage alignment per occupancy forecasting with consistent awareness from agent-wise motion predictions. Second, we propose a game-theoretic motion predictor, GTFormer, to model the interactive future among individual agents with their joint predictive awareness. Third, hybrid prediction patterns are concurrently integrated with Ego Planner and optimized by prediction guidance. HPP achieves state-of-the-art performance on the nuScenes dataset, demonstrating superior accuracy and consistency for end-to-end paradigms in prediction and planning. Moreover, we test the long-term open-loop and closed-loop performance of HPP on the Waymo Open Motion Dataset and CARLA benchmark, surpassing other integrated prediction and planning pipelines with enhanced accuracy and compatibility. Project website: {\href{https://georgeliu233.github.io/HPP}{https://georgeliu233.github.io/HPP}}
\end{abstract}


%%%%%%%%%%%%%%%%%%%%%%%%%%%%%%%%%%%%%%%%%%%%%%%%%%%%%%%%%%%%%%%%%%%%%%%%%%%%%%%%
%\linenumbers
\section{Introduction}
\label{sec1}
\begin{figure}[tp]
    \centering
    \includegraphics[width=\linewidth]{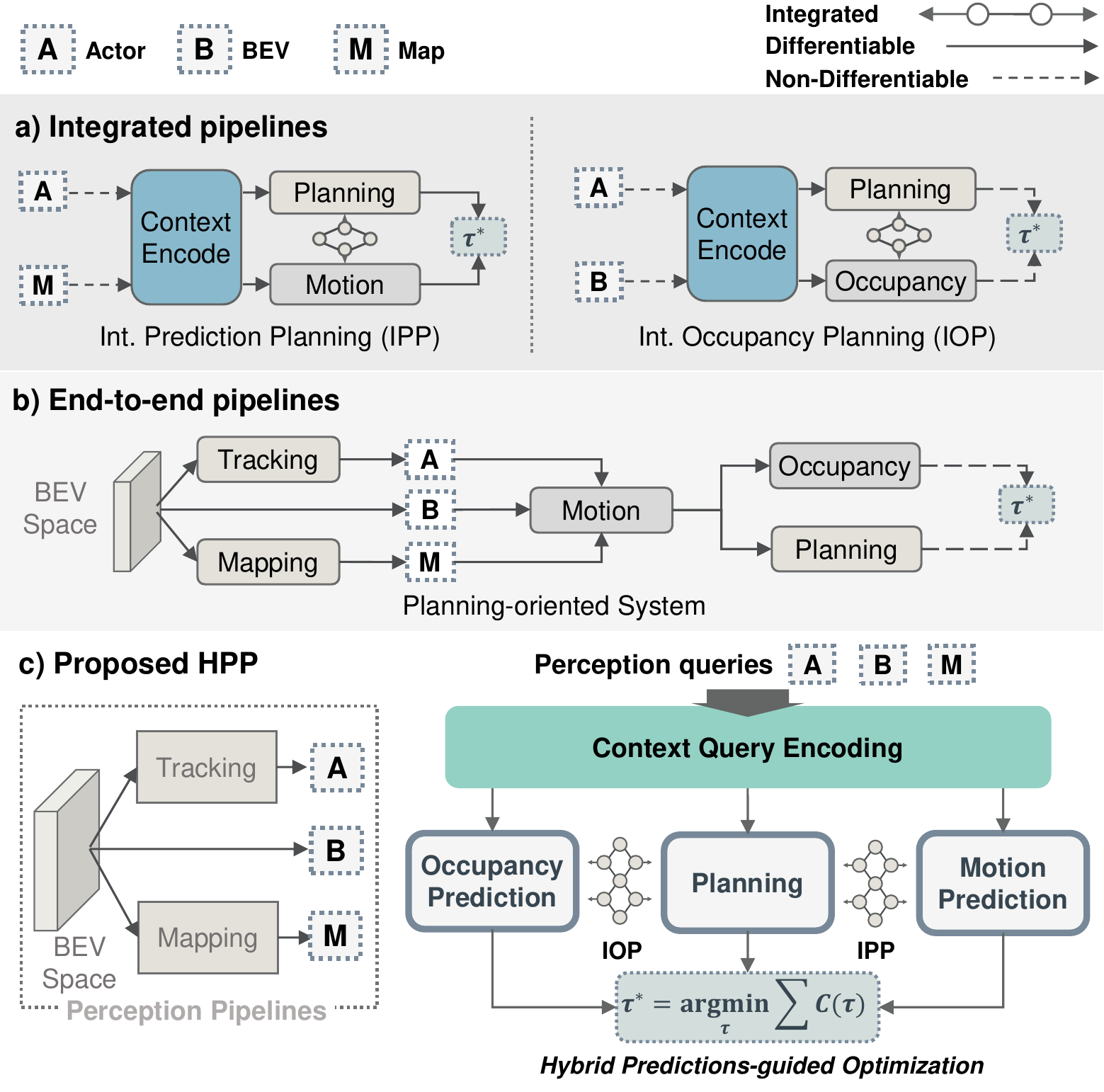}
    \caption{Generic learning-based pipelines in autonomous driving. a) Integrated pipelines learn planning jointly with motion prediction (IPP) or occupancy prediction (IOP) with coupled networks. b) End-to-end pipelines directly map from raw sensor inputs to planning. c) Our proposed HPP establishes a coupled planning system with hybrid-prediction integration and optimization.}
    \label{fig1}
    % \vspace{-0.1cm}
\end{figure}

\IEEEPARstart{A}{utonomous} driving systems (ADS) have made significant progress in perception, prediction, and planning, thanks to the advancement of learning paradigms \cite{chen2022milestones}. However, the performance growth of these independent tasks has come to a halt, prompting a reconsideration of modular design optimization \cite{hu2023planning,huang2022differentiable,hu2022st,jiang2023vad}. Fueled by inherent interactions among autonomous vehicles and traffic participants, recent research has placed significant emphasis on the integration of prediction and planning tasks \cite{hagedorn2023rethinking}. They seek to achieve concurrent advancements in both prediction and planning .

Integrated pipelines (Fig. \ref{fig1}a) generally connect planning and singular prediction modules involving agent-wise motion trajectories (IPP) \cite{huang2022differentiable,mo2022multi,pini2023safe} or whole-scene occupancy probabilities (IOP) \cite{liu2023occupancy,hu2023imitation,bansal2018chauffeurnet}. However, relying on single prediction format inevitably confronts respective shortfalls. Specifically, motion prediction models continuous temporal trajectories tracked for each agent, but encounters inconsistency and exponential cost in joint social patterns from marginal agents \cite{mozaffari2020deep}. Conversely, exclusive occupancy accurately predicts aligned joint patterns of whole scene agents under bird's eye view (BEV) perception, but the loss of agent-wise tractability leads to temporal conflicts and omission risks of critical agents \cite{kim2022stopnet}. This characteristic highlights the complementarity of motion and occupancy prediction. 
The inconsistency arising in the singular prediction pattern presents challenges for both IPP and IOP, resulting in incompatible planning learned collectively with misaligned predictions.

% integrates respective feedback from motion forecasting to planning, collectively learning interactive futures. However, IPP faces challenges due to exponential cost and social conflicts resulting from joint predictive patterns from marginal agents in planning with mutual awareness. To address this, integrated occupancy prediction and planning (IOP) uses dense probabilistic occupancy grids for one-shot joint predictions of arbitrary agents . Nonetheless, joint patterns by occupancy prediction lose marginal agent-wise tractability, causing intractable pixels and missing agents under assumptions for perfect perception. This also creates a domain gap in current integrated pipelines towards full-stack developments for ADS.

The limitations of integrated systems have spurred a renewed focus on end-to-end pipelines (Fig. \ref{fig1}b) \cite{chen2023end}. Planning-oriented systems employ a well-organized modular design to address plan guidance under a unified BEV geometry and query-based intermediate interactions \cite{hu2023planning,jiang2023vad,ye2023fusionad}. While this approach proves planning superiority and consider a hybrid format of predictions and planning, it falls short in addressing potential conflicts among different predictions.  Furthermore, the absence of interactive co-design across predictions and planning results in passive maneuvers and hinders the system learning for social compatibility. The integration of different modules gives rise to three major challenges: inherent trade-offs, including inconsistency and omission risks with sole prediction; discrepancies between joint and marginal patterns in predictions, limiting accuracy and impeding the learning of safe and naturalistic driving maneuvers; and the absence of interactive co-design for compatible planning, considering dependencies of hybrid prediction.

% The constraints of integrated systems have spurred a renewed emphasis on end-to-end pipelines. Planning-oriented systems employ a well-organized modular design to address plan guidance under a unified bird's eye view (BEV) geometry and query-based intermediate interactions. While this approach proves effective in handling a hybrid format of predictions and planning, it falls short in addressing potential conflicts among different predictions. Furthermore, the absence of interactive co-design across predictions and planning results in passive maneuvers and hinders social compatibility during the learning process. The integration of different modules gives rise to three major challenges: inherent trade-offs, including inconsistency and omission risks with sole prediction; discrepancies between joint and marginal patterns in predictions, limiting accuracy and impeding the learning of safe and naturalistic driving maneuvers; and the absence of interactive co-design for compatible planning, considering the dual dependencies of hybrid prediction.

%Two mainstream approaches have emerged, namely semi-modular systems and planning-oriented systems \cite{hu2023planning}. Semi-modular systems use multi-task learning (MTL) that sequentially maps raw sensor data to ego plan, but accumulated errors and diminishing geometry hinder safe and interpretable learning \cite{casas2021mp3,hu2022st,akan2022stretchbev}.

To tackle the aforementioned challenges, we propose a novel integrated pipeline (Fig. \ref{fig1}c) named \textbf{H}ybrid-\textbf{P}rediction integrated \textbf{P}lanning (\textbf{HPP}) that optimizes prediction and planning in a co-design process. By combining IPP and IOP, HPP delivers consistent planning while ensuring that hybrid prediction inform each other consistently. To achieve this, HPP leverages Transformer-based queries to channel and aggregate interactions between modules. Additionally, we have developed a refinement process that guides safe planning interacting with hybrid prediction. Specifically, we propose MS-OccFormer to perform marginal-conditioned joint occupancy prediction that aligns and refines consistently with marginal motion prediction. We leverage GTFormer, which is inspired by simulating game-theoretic iterative reasoning of marginal motion prediction with ego vehicle and joint occupancy. HPP concurrently models fine-grained interactions of integrated hybrid prediction patterns in Ego Planner, where we eventually devise a hybrid-prediction-guided refinement mechanism to facilitate safe and compatible planning. The main contributions are listed below:

\begin{enumerate}
\item We propose HPP, a modular co-design optimization ADS paradigm, that consistently interacts among marginal and joint prediction patterns with planning.

\item We introduce MS-OccFormer, a model that predicts joint occupancy patterns in BEV geometry while being aware of the marginal predictions. We also present GTFormer, a model that performs game-theoretic reasoning among marginal motion predictions in coordination with both planning and occupancy prediction. The ego planner is devised under interactive guidance by hybrid prediction.

\item HPP is tested on multiple large-scale real-world benchmarks, and extensive testing results demonstrate its state-of-the-art performance in terms of accuracy, safety, and consistency in prediction and planning.
\end{enumerate}

The remainder of this paper is organized as follows: Sec. \ref{sec2} reviews integrated prediction and planning pipelines in ADS. Sec. \ref{sec3} formulate the methodology of HPP. Comprehensive benchmarks and discussions are presented in Sec. \ref{sec4}. Finally, Sec. \ref{sec5} concludes the whole work.  

 % Another inherent problem is the ignorance in considering planning behavior causing incompatibility in planner guidance.
\section{Related Work}
\label{sec2}
\subsection{Predictions and Planning in ADS}
Prediction and planning modules in the conventional ADS model are separate. Predictions typically define evolving transitions as conditions for safe planning. Learning-based prediction models excel in modeling interactions among diverse agents and scene contexts \cite{mozaffari2020deep}. Categorized by representations, sparse predictions forecast multi-agent trajectories (MATP) along with detected participants. Leveraging Transformer \cite{jia2023hdgt,huang2022multi} or GNNs \cite{mo2022multi, mo2023map} in constructing the social interaction graph \cite{mo2022multi} or recurrent refining \cite{shi2022motion}, MATP filters multi-agent predictions in scoring combinations of marginal ones for each agent. While achieving agent-wise accuracy, MATP introduces exponential computations and trajectory-wise inconsistency. Dense predictions directly estimate the future distribution of agents jointly from ego-centered occupancy \cite{liu2023occupancy,hu2022hope,huang2022vectorflow}. A notable issue is the loss of agent-wise tractability. Enhancements via trajectories of heatmap sampling \cite{gilles2021thomas} or joint trajectory learning \cite{kamenev2022predictionnet} exhibit similar consistency issues. In ADS planning, various approaches are well-founded including sampling \cite{huang2023conditional}, optimization \cite{hang2020integrated}, and learning-based techniques via imitation learning \cite{xu2023bits} or reinforcement learning \cite{liu2022augmenting}. Still, achieving safe and socially compatible driving maneuvers in planning requires interactive awareness and safety guarantees derived from planning-compliant predictions. Furthermore, accumulated errors from detached predictions and planning underscore the need for developing integrated ADS.

\subsection{Integrated Predictions and Planning in ADS}
Navigating dense and interactive traffic requires integrated ADS, which models the simultaneous driving behaviors among social agents and the autonomous vehicle. Intuitive thought is to stack ego vehicles collectively with social agents' predictions and learn unified trajectories. All interactions are implicitly modeled by Transformer-based IPP \cite{pini2023safe, renz2022plant} or post-processed occupancy prediction\cite{kamenev2022predictionnet} under the IOP pipeline. To consider the evolving interactive behaviors between predictions and planning, conditional methods model the behavioral response individually from ego vehicle to social agents predictions \cite{rhinehart2019precog}. Conditional motion predictions are then integrated into planning reschedules \cite{huang2023conditional, huang2023learning} or modeled as non-cooperative games \cite{espinoza2022deep}. Still, these one-way interactions bypass the mutual consistency with all agents. Moreover, iterative bi-level optimization \cite{burger2022interaction} significantly slows down learning. Obviated from agent-wise conflicts, hierarchical game-theoretic approaches model the iterative reasoning process \cite{wang2022social}, updating mutual behaviors for all agents simultaneously \cite{huang2023gameformer}. Yet, uniformed agent-wise reasoning lacks specifications between predictions and planning, which should be target-driven. In HPP, we integrate reasoning by introducing an agent-conditioned occupancy to modulate joint behavior in social agents. and an Ego Planner for interactive planning. This integrated co-design enables learning both predictions and planning with flexibility while maintaining mutual awareness by iterative reasoning.

\begin{figure*}[t]
    \centering
    \includegraphics[width=\linewidth]{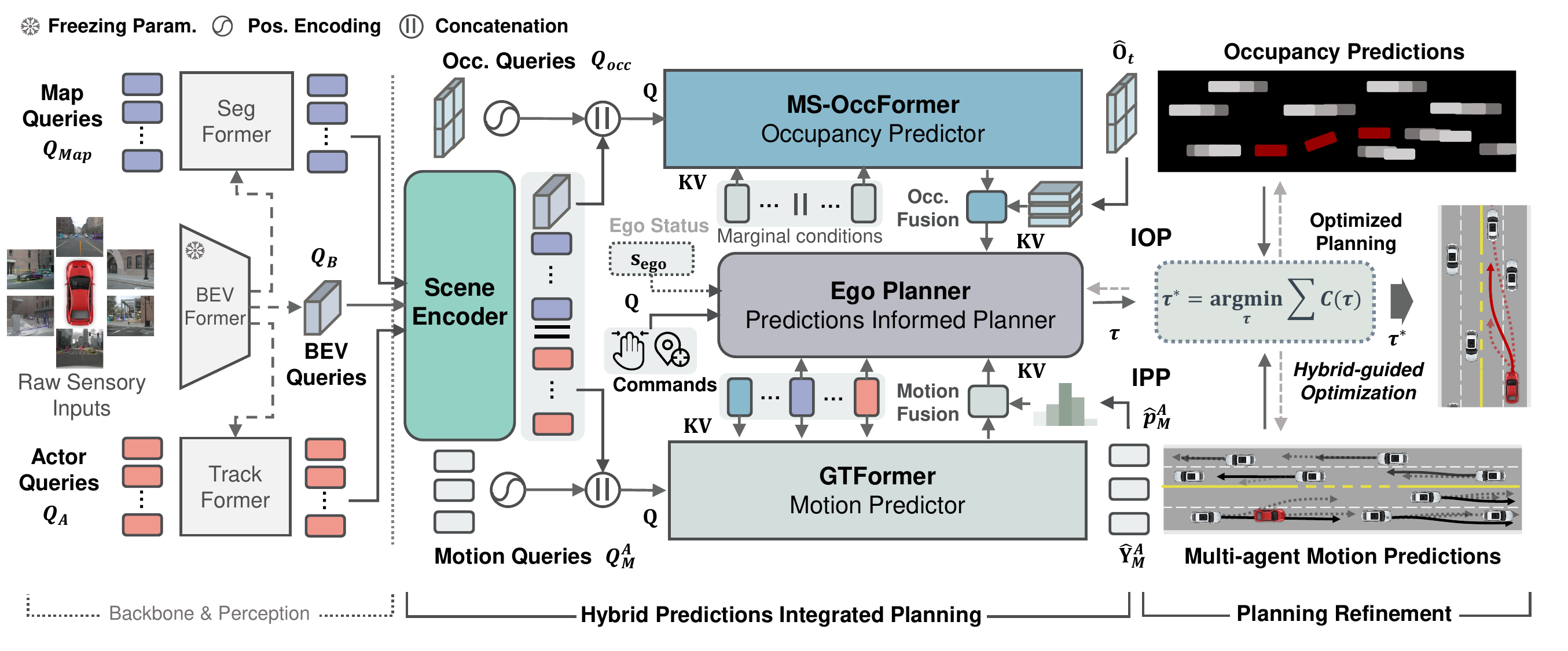}
    \caption{Systematic overview of the proposed \emph{Hybrid-Prediction integrated Planning} (\textbf{HPP}) framework. HPP is established upon query-based co-design optimization of interactive planning with hybrid prediction integration (IPP and IOP), informed by BEV perceptions. With encoded perception scene context $Q_{Map}, Q_B, Q_A$, HPP provides prediction and planning co-design in three-fold. Joint occupancy prediction $\hat{\textbf{O}}$ is iteratively refined in \textbf{MS-OccFormer}, sharing mutual consistency over marginal motion prediction $\hat{\textbf{Y}}$ in \textbf{GTFormer}. GTFormer performs interactive reasoning between marginal prediction and planning. Reasoned outcomes and ego features are then served to query hybrid prediction-aware planning $\tau$ in \textbf{Ego Planner}. Eventually, optimizations are scheduled to refine planning $\tau^*$ with hybrid-prediction guidance.}
    \vspace{-0.4cm}
    \label{fig2}
\end{figure*}

Recent paradigms focus on leveraging output responses for predictions to enhance planning guarantees. Adversarial objectives are employed between joint motion predictions and planning, considering likely \cite{huang2022differentiable, karkus2023diffstack}  or safety-critical patterns \cite{hanselmann2022king} for mutual differentiable optimizations. However, similar consistency issues faced by MATP pose additional optimization challenges. Some methods predict dense occupancy jointly as a potential cost for planning guidance \cite{bansal2018chauffeurnet, liu2023occupancy, hu2023imitation}. However, intractability issues introduce risks and require laborious filtering. An additional challenge is the lack of integrated modeling leads to inconsistency and limits the interactive process only in optimization. In HPP, integration is considered concurrently for ADS co-design and output optimizations. Through flexible and consistent hybrid prediction and planning via system co-design, hybrid prediction are jointly utilized to refine planning, enhancing safety consistently.

\subsection{End-to-end Systems and LLMs for Autonomous Driving}
End-to-end methods consider a direct mapping from raw sensors to prediction and planning under perception understanding \cite{chen2023end}. A typical system is serialized by modules and learns jointly with each objective \cite{casas2021mp3, hu2022st}. However, modular errors are accumulated and geometry is hindered by misaligned predictions and planning. Thus, planning is sampled in trajectory retrieval by predictions \cite{liang2020pnpnet}. Prominence in BEV perception \cite{li2022bevformer, zhang2022beverse,akan2022stretchbev} enables modular integration and learning under unified BEV geometry \cite{li2023delving}. This further prompts a planning-oriented system, which organizes and serves all intermediate modules targeting planning under visual \cite{hu2023planning, ye2023fusionad, jia2023driveadapter} or vectorized \cite{jiang2023vad} perceptions. Query-based design channel and propagate modular integration. This prompts recent works incorporating large language models (LLMs) as the motion planner \cite{mao2023gpt} or routing agent \cite{mao2023language}. Still, the current end-to-end system focuses more on integration with perception that aligns predictions and planning, leaving paradigm discussions incorporating LLMs \cite{sima2023drivelm}. In HPP, we put more emphasis on modular co-design optimization, learning interactive hybrid prediction, and planning informed by perception under query-based integration.

\section{Hybrid-prediction Integrated Planning}
\label{sec3}
The overview of our proposed \textbf{HPP} is shown in Fig.\ref{fig2}, which is defined in Sec. \ref{sub1} upon query-based modular co-design and optimization of ADS. In the upcoming sections, informed by BEV perception pipelines exhibited in Sec. \ref{sub2}, HPP manages the framework co-design from \textbf{MS-OccFormer} in Sec. \ref{sub3} for joint occupancy prediction, sharing mutual consistency with motion prediction in \textbf{GTFormer}. Here we elaborate its hierarchical reasoning model for interactive prediction and planning in Sec. \ref{sub4}. Followed by hybrid prediction-aware \textbf{Ego Planner} in Sec. \ref{sub5}, we frame the learning and optimization process for HPP reckoning hybrid prediction guidance in Sec. \ref{sub6}.
\subsection{Problem Formulation}
\label{sub1}
As shown in Fig. \ref{fig2}, HPP focuses on addressing integrated predictions and planning challenges for ADS. Informed by perception under BEV geometry, this learning-based system is founded by modular co-design and optimization. Modularized by Transformer basis, HPP leverages categorical queries $\mathbf{Q}$ to aggregate modular outputs and channel interactions as keys and values $\mathbf{K}, \mathbf{V}$ by multi-head attention mechanism. 

\begin{figure*}[t]
    \centering
    \includegraphics[width=\linewidth]{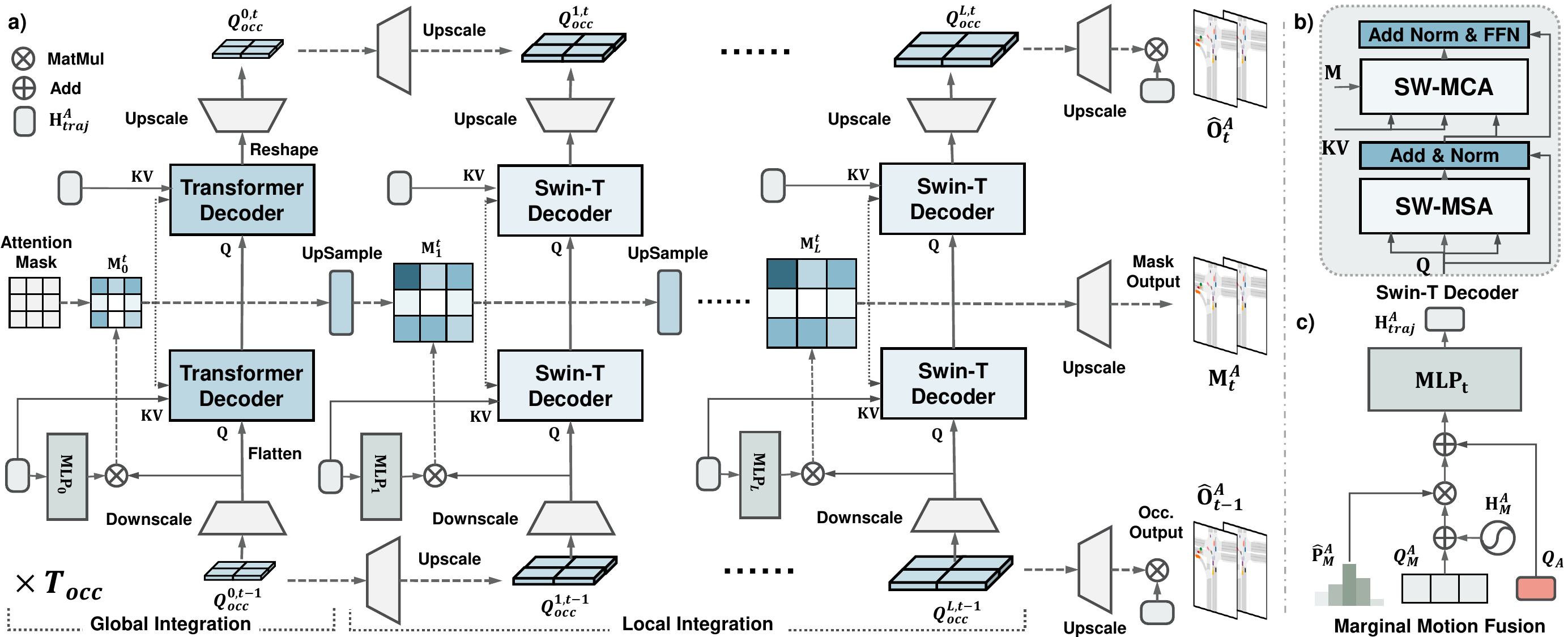}
    \caption{Generic learning framework in MS-OccFormer: a) A single block of multi-scale marginal-conditioned occupancy predictor. Joint occupancy $\hat{\mathbf{O}}$ is consistently integrated with marginal prediction features $\mathbf{H}^A_{traj}$ through global interactions and local refinements, and guided by iteratively updated learnable attention mask $\mathbf{M}$; b) A Swin-T decoder for local interactions through shifted-window cross attention; c) Agent-wise fusion for marginal prediction features.}
    \vspace{-0.3cm}
    \label{fig3}
\end{figure*}

Given multi-view image inputs, the co-design of HPP begins with the perception backbone for BEV features $\mathcal{B}$. BEV perception further defines an ego-centered area of $H\times W$ with map $\mathcal{S}$ and $N_A$ detected agents $\mathcal{A}\in A_{0:N_A-1}$ at current timestep $t=0$, where $A_0$ denotes the ego vehicle. Based on scene context $\mathbf{X}=\{\mathcal{B},\mathcal{S}, \mathcal{A}\}$, HPP aims to learn and optimize concurrently for hybrid prediction and planning modules. Specifically, over future horizon $T$, and set of queries $\mathbf{Q}$, joint predictions are defined as per-step occupancy probabilities $\mathbf{\hat{O}}_{1:T}=\{o^{h,w}_t|o^{h,w}\in[0,1]^{H\times W}\}_{t=1}^{T}$ for all neighbor agents. Simultaneously, the marginal future motions for all agents are defined by $\hat{\mathbf{Y}}^A_M=\{(\hat{\mathbf{y}}^{1:T,i}_m, \hat{\mathbf{p}}^i_m)|m\in[1,M]\}^{N_A-1}_{i=0}$ denoting multi-modal future trajectories $\hat{\mathbf{y}}\in \mathbb{R}^{N_A \times M \times T \times 2}$ and probability $\hat{\mathbf{p}}\in \mathbb{R}^{N_A \times M}$ considering $M$ modes of uncertainty. hybrid prediction-aware planning $\tau_{1:T}\in \mathbb{R}^{T \times 2}$ is queried by plan context $\mathcal{E}$. Subsequently, HPP formulates a collective objective $\mathcal{L}$ for modular learning, and cost criteria $\mathcal{C}$ for optimized hybrid prediction guided planning $\tau^*\in \mathbb{R}^{T \times 2}$. In general, HPP is formulated as:

\begin{equation}
\label{e1}
\begin{aligned}
    \mathbf{\hat{Y}}^A_M, \mathbf{\hat{O}}_{1:T}, \tau_{1:T} &= f(\mathbf{X}, \mathbf{Q}, \mathcal{E} | \theta), \\
    \tau^*=\mathop{\arg\min}_{\tau}& \ \mathcal{C}(\tau, \mathbf{X}, \mathbf{\hat{Y}}, \mathbf{\hat{O}}),
\end{aligned}
\end{equation}
where $f$ denotes the co-design of HPP, and $\theta$ is the model parameters.
Specific co-design and formations for each module in HPP are illustrated in the following sections.

\subsection{Perception Scene Encoding}
\label{sub2}

Perception pipelines in HPP aim to capture scene context $\mathbf{X}$ with raw image inputs under BEV geometry. Scene contexts are then jointly encoded to catch their global relations.

\subsubsection{BEV Perception} In HPP, multi-view image features $\mathbf{I}\in \mathbb{R}^{6 \times H_{in} \times W_{in} \times C}$ are  extracted via shared backbones \cite{he2016deep} from raw image inputs. We leverage BEV encoder \cite{li2022bevformer} to transform $\mathbf{I}$ into BEV feature $\mathcal{S}$ through recurrent top-down BEV queries $Q_B \in \mathbb{R}^{H \times W \times D}$, where $D$ denotes the hidden dimensions. Founded on BEV backbones, we utilize two DETR-like perception decoders \cite{hu2023planning} in extracting scene context features for agent $\mathcal{A}$ and map $\mathcal{S}$ from $Q_B$, following $N_A$ agent queries $Q_A \in \mathbb{R}^{N_A \times D}$ and $N_M$ map queries $Q_{Map} \in \mathbb{R}^{N_M \times D}$. Note that HPP focuses on integrated prediction and planning. Therefore, it is expected better results for HPP using more advanced BEV perception units.

\subsubsection{Scene Encoding} to model the global interactions between scene elements with BEV perceptions, we inherit from our previous work \cite{liu2023multi,liu2023occupancy} to gather and encode separate visual and vectorized scene context features. Specifically, visual features are encoded by BEV queries $Q_B$, and the scene features for map and agent are concatenated and encoded as $Q_s=[Q_{Map};Q_A] \in \mathbb{R}^{(N_A+N_M) \times D}$, where $[\cdot;\cdot]$ denotes concatenations. Encoded results $\mathbf{X}=\{Q_B, Q_{Map}, Q_A\}$ are then served as input for HPP co-design integrating predictions and planning.
%With detached $Q_B$ from pretrained BEV backbone, we inherit from the warm-up stage that train two perception decoders primarily with joint perception objectives $\mathcal{L}_{per}$ comprising detection, tracking and segmentation for actor and map feature.

\subsection{Ms-OccFormer}
\label{sub3}
HPP formulates joint predictions as occupancy $\mathbf{O}_{1:T}$ consistently with BEV geometry in perception. To further tackle the consistency challenges between hybrid prediction, in HPP we propose MS-OccFormer spotlights twp aspects, i.e. Marginal-conditioned occupancy that defines tractable predictions, and Multi-scale prediction-wise integration that deals with the interactive alignments with different granularity. Illustrated in Fig. \ref{fig3}, MS-OccFormer utilizes a streaming-based pipeline to roll out the future horizon $T$
, decoding per-step occupancy prediction based on $L$ levels integration of succeeded step features.

\subsubsection{Modular Queries} we leverage occupancy queries $Q_{occ}\in \mathbb{R}^{H \times W \times D}$ in  multi-scale aggregating for  positional and perception features: $Q_{occ} = \operatorname{MLP}([\operatorname{PE}({I}_{B}); Q_B])$. Positional grids ${I}_{B}\in\mathbb{R}^{ \times H \times W \times 2}$ are encoded using sinusoidal $\operatorname{PE}(\cdot)$ and transformed by multi-layer perceptron (MLP). We further downsample $Q_{occ}$ under $L$ levels $\{Q^{l,0}_{occ}\in\mathbb{R}^{\frac{H}{2^{l}}\times\frac{W}{2^{l}} \times D}\}^{0}_{l=L}$ to recurrently query multi-scale interactions.

\subsubsection{Marginal Dependencies} \label{margin_dep} To fully extract the interactive marginal prediction features, we conduct an agent-wise fusion (see Fig. \ref{fig3}c) that leverages the marginal future $\hat{\mathbf{Y}}^A_M$ outputs from GTFormer (Sec. \ref{sub4}). Multi-modal motion features $Q^A_M\in \mathbb{R}^{N_A\times M\times D}$ are fused with marginal features $\mathbf{H}^A_M=\operatorname{MLP}(\operatorname{PE}(\mathbf{\hat{y}}))$ and projected by each horizon: 
\begin{equation}
    \mathbf{H}^A_{traj} = \operatorname{MLP}_{1:T}(\mathbf{\hat{p}}(Q^A_M + \mathbf{H}^A_M) + Q_A),
    \label{equ2}
\end{equation}
where $\mathbf{H}^A_{traj}\in \mathbb{R}^{T \times N_A \times D}$ denotes the marginal features.

\subsubsection{Marginal-conditioned Occupancy} The primary challenge in formulating $\mathbf{O}_{1:T} \in \mathbb{R}^{T \times H \times W}$ lies in the intractability with marginal predictions that cause joint inconsistencies. Inspired by instance-level occupancy $\mathbf{O}^A_{1:T} \in \mathbb{R}^{T  \times H \times W \times N_A}$ \cite{hu2023planning} and conditional methods \cite{huang2023conditional}, we propose the marginal-conditioned occupancy prediction task. This models the consistent joint occupancy $p(\mathbf{O}^A_{1:T}|\mathbf{Y}^A_M,\mathbf{X})$ over agent-wise marginal predictions. To associate uncertainty and mutual interactions, given final joint decoding features $Q^L_{occ}$ and marginal features $\mathbf{H}^A_{traj}$, the marginal-conditioned occupancy will be eventually modeled by dot products:
\begin{equation}
    \mathbf{O}^A_{1:T} = \sigma(Q^L_{occ}\cdot \operatorname{MLP}(\mathbf{H}^A_{traj})^T), 
    \label{equ3}
\end{equation}
where $\sigma$ denotes the sigmoid function for per-grid probabilities. The original task can be then transformed back $\mathbf{O}_{1:T}=\max_{A}\mathbf{O}^A_{1:T}$ for co-design of other HPP modules:

\subsubsection{Multi-scale Prediction-wise Integration}\label{multiscale} aims to iteratively align multi-scale interaction features between hybrid prediction in decoding $\mathbf{O}^A_{1:T}$.In Fig. \ref{fig3}a, multi-scale succeeded occupancy features \(\{Q^{l, t-1}_{occ}\}^L_{l=1}\) query aligned marginal features by attentions at different granularities from two-stage Transformer decoders. 

The global integration stage leverages the vanilla Transformer decoders to perform per-grid interactions from flattened high-level joint features $Q^{L, t-1}_{occ}$ with marginal ones. Subsequently, with the upscaling of occupancy features $\{Q^{l, t-1}_{occ}\}^{L-1}_{l=1}$, the local integration stage focuses on capturing consistency from partial joint behaviors with marginal features. This motivates us to design shift-window multi-head cross-attention (SW-MCA), inspired by SW-MSA in \cite{liu2021swin}. As depicted in Fig. \ref{fig3}b, we employ the rolling process to simultaneously capture local interactions under shifted windows attention.

To ensure interactive consistency across multi-scale integration, we devise a learnable attention mask $\mathbf{M}^l_{1:T}\in\mathbb{R}^{T\times \frac{H}{2^{L-l}}\times\frac{W}{2^{L-l}}\times N_A}$ for Transformer decoder that iteratively refines upon interaction results from the previous scale. This aligns the attention modeling based on the previous results Shown in Fig. \ref{fig3}a, for each level, the attention mask gets updated with agent-conditioned occupancy on the current scale level:
\begin{equation}
    \mathbf{\hat{M}}^l = \sigma(Q^l_{occ}\cdot \operatorname{MLP}_l(\mathbf{H}^A_{traj})^T).
\end{equation}
The attention masks are then iteratively updated following:
\begin{equation}
    \mathbf{M}^l = \lambda_{m}\operatorname{Upsample}(\mathbf{M}^{l-1}) + (1- \lambda_{m})\mathbf{\hat{M}}^l,
\end{equation}
where $\lambda_{m}=0.5$ is the update factor. In general, given Transformer decoder at certain stage as $\operatorname{Trans}$, the prediction-wise integration under scale $l$ of timestep $t$ is defined as:
\begin{equation}
   Q^{l, t}_{occ} = \operatorname{Trans}(q=Q^{l, t-1}_{occ}, k,v=\mathbf{H}^{A,t}_{traj},m=\mathbf{M}^l_t).
\end{equation}
Output joint occupancy features $Q^{L}_{occ}$ will be eventually fused via Equ. \ref{equ3} for conditioned occupancy predictions $\mathbf{\hat{O}}^A_{1:T}$.

% \begin{equation}
% \operatorname{SW-MCA} = \operatorname{SW-MSA}(\operatorname{roll}(\mathbf{Q}), \mathbf{K}, \mathbf{V}, \operatorname{roll}(\mathbf{M})\cdot\mathbf{M}_w)
% \end{equation}

\begin{figure*}[t]
    \centering
    \includegraphics[width=\linewidth]{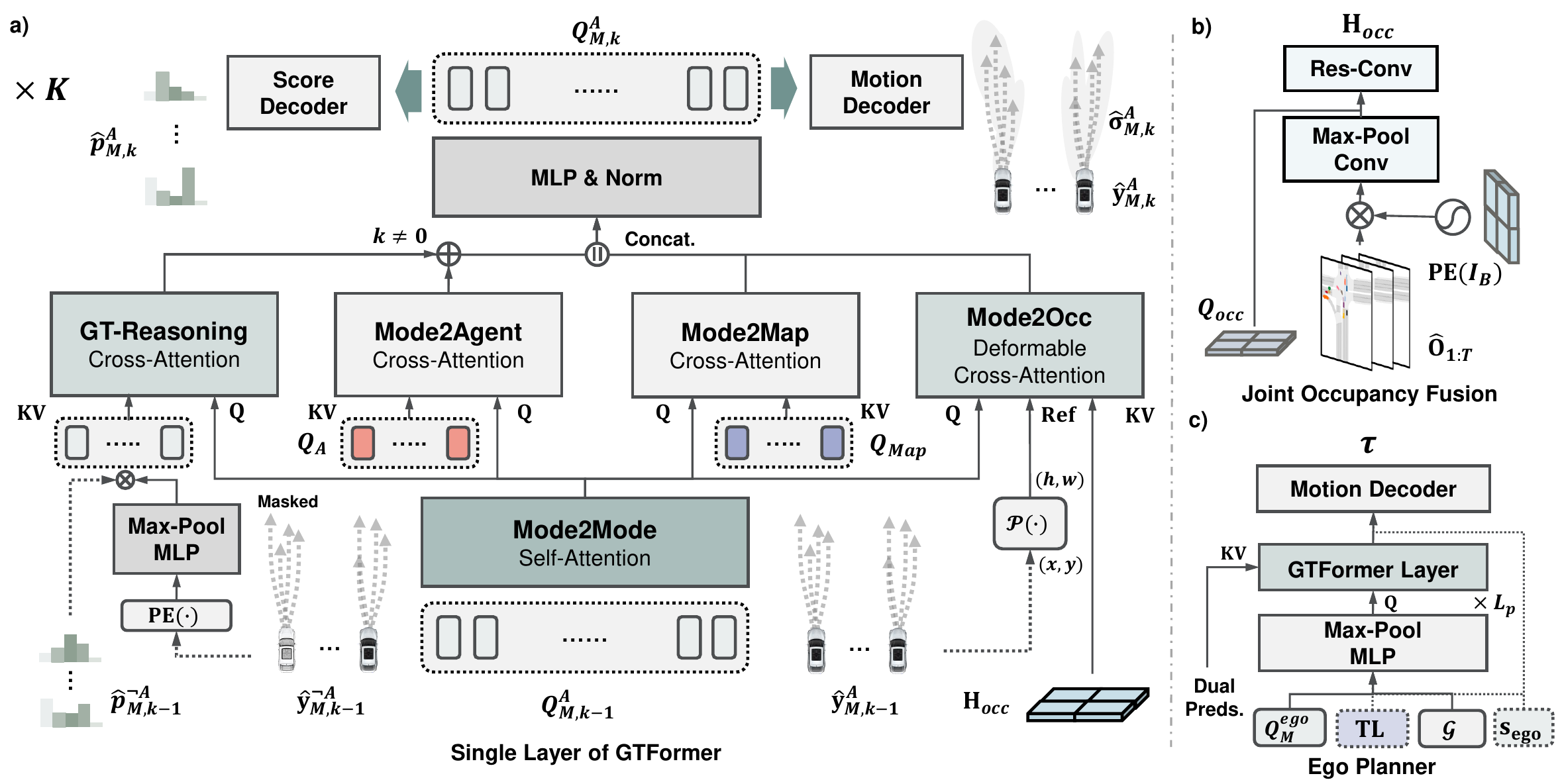}
    \caption{a) A single-step reasoning layer of GTFormer. Level-$K$ reasoning queries interactive behaviors hierarchically for all agents $Q^A_M$ in predictions and planning. Meanwhile, it considers interactions with scene context $Q_A, Q_{Map}$ and joint predictive BEV features $\mathbf{H}_{occ}$; b) Occupancy fusion for joint prediction features; c) hybrid-prediction aware Ego Planner conditioned on plan context.} 
    \vspace{-0.3cm}
    \label{fig4}
\end{figure*}

\subsection{GTFormer}
\label{sub4}
In HPP, ensuring interactive consistency between motion predictions and planning involves the introduction of GTFormer. As depicted in Fig. \ref{fig4}, GTFormer formulates future interactive behaviors as a game-theoretic reasoning process, simulating hierarchical reasoning through $K$-layers stacked Transformer decoders. Beyond the uniform agent-wise reasoning model of our previous work \cite{huang2023gameformer}, GTFormer collaborates with MS-OccFormer (Sec. \ref{sub3}), modeling occupancy interactions to modulate joint future behaviors at each level of reasoning.

\subsubsection{Modular Queries} We leverage motion queries $Q^A_M\in \mathbb{R}^{N_A\times M\times D}$ to initialize agent-wise interactive reasoning with multi-modal motion intentions ${I}_A \in \mathbb{R}^{N_A\times M\times 2}$ and agent perception features by:
$Q^A_M = \operatorname{MLP}([\operatorname{PE}({I}_{A}); Q_A])$. 

\subsubsection{Joint Dependencies} \label{joint_dep} We transform joint occupancy predictions $\hat{\mathbf{O}}_{1:T}\in\ \mathbb{R}^{T\times H \times W}$ with positional features $I_B$ to align with continuous geometry for predictions and planning. Shown in Fig. \ref{fig4}b, we conduct a multiplication with max-pooling to encode occupancy positional features with BEV semantics: 
\begin{equation}
   \mathbf{H}_{occ} = \operatorname{ResConv}( \mathop{\max}_{T}\operatorname{Conv}(\operatorname{PE}(I_B)\hat{\mathbf{O}})+ Q_{occ}),
\end{equation}
where \(\operatorname{Conv}\) and \(\operatorname{ResConv}\) represent convolutional projections and residual layers, respectively. The joint features \(\mathbf{H}_{occ}\in \mathbb{R}^{H\times W\times D}\) are then interacted with all agents.

\subsubsection{Game-Theoretic Transformer Layer} We employ level-k games to model the GTFormer layer (see Fig. \ref{fig4}a) for future interactive behaviors of all agents. Inherited from our previous work \cite{huang2023gameformer}, we denote agent-wise multi-modal predictions and planning $\hat{\mathbf{Y}}^A_M$ as player policy $\{\pi_i\}^{N_A-1}_{i=0}$. For each player $i$, the reasoning process is defined as a $K$-times iterations of policy conditioned on the opponents $\neg i$ policy from last reasoning level: $\pi^k_i(\mathbf{Y}^i_M|\pi^{k-1}_{\neg i},\mathbf{X}), k\in[1,K-1]$. Specifically, the level-0 policy is reasoned independently: $\pi^0_i(\mathbf{Y}^i_M|\mathbf{X})$. We leverage the Gaussian mixture model (GMM) to outline the uncertainty for predictions and planning policy as:  
\begin{equation}
   \pi^k_i = \sum^{M}_{m=1}p(\mathbf{\hat{p}}^i_m|\pi^{k-1}_{\neg i}, \mathbf{X})\sum^{T}_{t=1}\phi(\mathbf{\hat{y}}^{t,i}_m, \mathbf{\hat{\sigma}}^{t,i}_m|\pi^{k-1}_{\neg i},\mathbf{X}),
   \label{equ8}
\end{equation}
where $\phi$ and $\mathbf{\sigma}^{t,i}_m \in \mathbb{R}^{N_A\times M\times T \times 2}$ represent the density function and variance, respectively for 2D Gaussian distributions.

Specifically, a single layer of GTFormer is shared by level-$k$ policy $\pi^k_{0:N_A-1}$ (Fig. \ref{fig4}a) across all agents. The $k^{th}$ layer starts with a multi-head self-attention (MHSA), named \textbf{Mode2Mode} for last level motion queries $Q^A_{M,k-1} \in \mathbb{R}^{N_A\times M \times D}$ to model future interactions across modalities. Four multi-head cross-attention (MHCA) are devised for $Q^A_{M,k-1}$ to aggregate the policy conditions separately for reasoning. (1) \textbf{GT-reasoning} interacts with last level components policies $\pi^{k-1}_{0:N_A-1}$. Fused by Equ. \ref{equ2} with $\mathbf{\hat{p}}^{\neg A}_{M,k-1}$ and $\mathbf{\hat{y}}^{\neg A}_{M,k-1}$, components features are filtered by future mask \cite{huang2023gameformer} to query future reasoning behaviors as $F^k_{GT}$.  (2) \textbf{Mode2Agent} and (3) \textbf{Mode2Map} outlines the scene context interactions of agents $Q_A$ and maps $Q_{Map}$ as $F^k_{A}$ and $F^k_{Map}$. (4) \textbf{Mode2Occ} leverage deformable MHCA (DCA) \cite{li2022bevformer} to model the future interactions with joint occupancy feature. $\mathbf{H}_{occ}$ is queried by A set of offsets $\Delta \in \mathbb{R}^{N_A\times M \times N_p \times 2}$ referenced by $\mathcal{P}(\mathbf{\hat{y}}^{A}_{M,k-1})$ transformed to pixel coordinate. $N_p$ referenced occupancy are then aggregated as $F^k_{occ}$ for each agent. In general, all interactive features are concatenated to update the motion query:
\begin{equation}
   Q^A_{M,k} = \operatorname{FFN}([F^k_{GT}+F^k_{A}; F^k_{Map}; F^k_{occ}]),
\end{equation}
where $\operatorname{FFN}(\cdot)$ denotes the feed-forward layers for residual MLPs and layer-norm. We omit the reasoning attention of $F^0_{GT}$ at $k=0$ for independent future policies. Eventually, updated motion queries $Q^A_{M,k}$ are passed through score and trajectory decoders for reasoned GMM policies of predictions and planning.

\subsection{Ego Planner}
\label{sub5}
HPP introduces the Ego Planner depicted in Fig. \ref{fig4}c. Ego Planner specifies planning-oriented conditions $\mathcal{E}$ based on the planning reasoning policy $\pi^{K-1}_0$. The co-design of HPP facilitates a hybrid-prediction aware Ego Planner that collaborates with GTFormer and MS-OccFormer, defined as $p(\tau|\pi^{K-1}_0, \hat{\mathbf{Y}}, \hat{\mathbf{O}}, \mathcal{E}, \mathbf{X})$. This is essential as the game-theoretic process in GTFormer (Sec. \ref{sub4}) simply models ego planning uniformly with other agents.

\subsubsection{Modular Queries} In the context of target-driven planning, we define plan queries \(Q_{\mathcal{E}}\in \mathbb{R}^{1 \times D}\) that integrate planning reasoning features \(Q^0_{M,K-1}\) using plan context \(\mathcal{E}\) through a max-pooling fusion by: $Q_{\mathcal{E}} = \mathop{\max}_{M}\operatorname{MLP}([Q^0_{M,K-1};\mathbf{H}_\mathcal{E}])$.
Here,  \(\mathbf{H}_\mathcal{E}\in\mathbb{R}^{1 \times D}\) encompasses global target features \(\mathcal{G}\in \mathbb{R}^{1 \times D}\) for encoded navigation commands or coordinates, traffic light features \(\operatorname{TL}\in \mathbb{R}^{1 \times D}\), and optional ego status \(\mathbf{s}_{\text{ego}}\in \mathbb{R}^{1 \times D}\) for ego speed and heading $\{v, \psi\}$.

\subsubsection{Hybrid Predictions Dependencies} We directly utilize a stack of \(L_p\) interaction Transformers parts in the GTFormer layer (Fig. \ref{fig4}a) to model future joint interactions \(\hat{\mathbf{O}}\) and marginal interactions \(\hat{\mathbf{Y}}\), informed by reasoned planning features and target-conditioned planning queries \(Q_\mathcal{E}\). The GT-Reasoning module incorporates hybrid prediction awareness for marginal motion interactions \(\mathbf{\hat{Y}}^{\neg A}_{M,K-1}\), while the Mode2Occ module handles joint occupancy interactions \(\mathbf{\hat{O}}_{1:T}\). By aggregating interactive future behaviors from hybrid prediction, plan queries pass through an identical motion decoder, resulting in a refined planning trajectory \(\tau\in\mathbb{R}^{T \times 2}\).

\subsection{System Learning and Optimization Design}
\label{sub6}
We present a collaborative learning and optimization paradigm for the HPP system. With detached BEV backbones, the co-designed framework undergoes a two-stage training process: 1) warm-up learning of two perception decoders using perception objectives \(\mathcal{L}_{per}\) \cite{hu2023planning}; 2) end-to-end supervised learning using all modular objectives \(\mathcal{L}\) as:
\begin{equation}
  \mathcal{L} = \mathcal{L}_{per} + \mathcal{L}_{occ} + \mathcal{L}_{GT} + \mathcal{L}_{plan}.
\end{equation}
During inference, hybrid-prediction guided optimization refines the planning \(\tau^*\) by minimizing cost functions \(\mathcal{C}\). In the following, we elaborate on modular objectives, costs, and co-optimization strategies.

\subsubsection{Modular Objectives} For precise prediction of marginal-conditioned occupancy \(\mathbf{\hat{O}}^A_{1:T}\) in MS-OccFormer, we employ a combination of top-k BCE loss and Dice loss \cite{hu2023planning} jointly for \(\mathbf{\hat{O}}^A_{1:T}\) and $\mathbf{M}_{1:T}$, for balanced predictions of occupancy probabilities: \(\mathcal{L}_{occ} = \mathcal{L}_{\text{BCE}} + \lambda_{\text{Dice}}\mathcal{L}_{\text{Dice}}\), with \(\lambda_{\text{Dice}}=5\).

To capture the hierarchical reasoning process for all agents in GTFormer, the min-max objectives for policy in level-$k$ consist of \(\mathcal{L}^k_{\text{IL}}\), minimizing imitative behaviors, and \(\mathcal{L}^k_{\text{col}}\), maximizing the interactive distance. The overall objective is defined as \(\mathcal{L}_{GT} = \sum^{K-1}_{k=0}(\mathcal{L}^k_{\text{IL}} + \lambda_{\text{col}}\mathcal{L}^k_{\text{col}})\). Here, \(\mathcal{L}^k_{\text{IL}}\) represents the negative log-likelihood (NLL) loss for reasoning policies \(\pi^{k}_{0:N_A-1}\), with the closest final displacement errors (FDE) for each agent as a positive mixture:
\begin{equation}
\mathcal{L}^k_{\text{IL}} = \sum_{i=0}^{N_A-1}\sum_{t=1}^{T}\sum_{m=1}^{M} \mathbf{1}(m=\hat{m}^i)\mathcal{L}_{\text{NLL}}(\hat{\mathbf{y}}^{t,i}_{m}, \mathbf{\sigma}^{t,i}_{m}, \hat{\mathbf{p}}^i_{m}),
\end{equation}
where $\hat{m}$ denotes the positive index, $\mathcal{L}_{\text{NLL}}$ is defined as:
\begin{equation}
\small
\mathcal{L}_{\text{NLL}} = \log \sigma_x + \log \sigma_y + \frac{1}{2} ((\frac{\Delta_x}{\sigma_x})^2 + ( \frac{\Delta_y}{\sigma_y})^2) - \log (p),
\end{equation}
Here $\sigma_x, \sigma_y \in \mathbf{\sigma}^{t,i}_{m}$ and $\Delta_{xy}=\mathbf{y}^{t,i}_{xy} - \hat{\mathbf{y}}^{t,i}_{m,xy}$. We leverage cross-entropy loss for $\hat{\mathbf{p}}^i_{m}$ in updating scoring. For interactive loss \(\mathcal{L}^k_{\text{col}}\), it is modeled by maximize L2 distance $\mathcal{D}$ for closest trajectories within $d_{\text{col}}$ under last level component policies:
\begin{equation}
\mathcal{L}^k_{\text{col}} = \sum_{i=0}^{N_A-1}\sum_{t=1}^{T}\sum_{m=1}^{M}\mathop{\max}_{j\neg i,M} \frac{\mathbf{1}(\mathcal{D}<d_{\text{col}})}{(1 + \mathcal{D}(\hat{\mathbf{y}}^{i,t,k}_{m},\hat{\mathbf{y}}^{j,t,k-1}_{M}))}.
\end{equation}

To perform refined planning $\tau$ in Ego Planner, it is learned with L2 distance $\mathcal{D}$: $\mathcal{L}_{plan} = \mathcal{D}(\tau,\mathbf{y}^{0})$.

\subsubsection{Cost Profiles} The cost function profiles encompass diverse cost terms $\{c_i\}$, taking into account various aspects of planning performance, categorized as: driving progress, comforts, adherence to rules \cite{liu2023occupancy}, and, crucially, safety. 
Importantly, the planning safety is defined by Gaussian potential fields incorporating joint $\hat{\mathbf{O}}$ and marginal $\hat{\mathbf{Y}}$ predictions. Motivated by complementary safety guidance by hybrid prediction, the explicit catalog is as follows:
\begin{equation}
    c^{\operatorname{safe}}_t = \sum_{\hat{\mathbf{O}}^{x,y}_t \in D_1} \phi(\tau_t, \hat{\textbf{O}}_t) + \sum^{N_A-1}_{i=1}\sum^M_{m=1}\sum_{\hat{\textbf{y}}^t\in D_2}\phi(\tau_t, \hat{\textbf{y}}^{t,i}_{m}).
\end{equation}
Here, \(\phi\) represents the Gaussian density functions as in Equ. \ref{equ8}. \(\hat{\mathbf{O}}^{x,y}_t = \mathcal{P}^{-1}(\hat{\mathbf{O}}^{h,w}_t)\) denotes the occupied coordinates, and \(\hat{\textbf{y}}^{t,i}_{m}\) is derived from reasoned results of \(\mathbf{\hat{Y}}^{A}_{M,K-1}\). Each potential field is subjected to masking by a distance threshold \(D_1=5, D_2=3\) towards planning trajectories.

\subsubsection{Optimization} The optimization for hybrid prediction-guided planning is defined as an open-loop optimization problem under finite horizons. The general formulation is as follows:
\begin{equation}
    \mathbf{u}^*=\mathop{\arg\min}_{\mathbf{u}} \frac{1}{2}\sum_i\Vert \omega_i c_i(\mathbf{u}, \mathbf{X}, \mathbf{\hat{Y}}, \mathbf{\hat{O}})\Vert^2,
\label{equ15}
\end{equation}
where $\mathbf{u}$ is the planning variable, and $\omega_i$ denotes the weight for cost function $c_i$. For generalized ADS, HPP solves this optimization problem according to different criteria:

 \textbf{Reference routes:} Suppose an accessible reference route \(\mathcal{I}\in \mathbb{R}^{L_{\text{ref}}\times d_\text{ref}}\) that is densely interpolated, HPP transforms all cost profiles under Frenet coordinates to alleviate optimization difficulties. Each reference point \(r \in \mathcal{I}\) is assigned with tangential and normal vectors: \([\Vec{t}_r, \Vec{n}_r]\). The Cartesian coordinate \(\Vec{y}=(x,y)\) can then be transformed to \(\Vec{r}=(s,d)\) via:
\begin{equation}
\Vec{y}(s(t),d(t))=\Vec{r}(s(t)) + d(t)\Vec{n}_r(s(t)).
\end{equation}

\textbf{Path planning}: 
This handles trajectories of lower frequency for the ego vehicle. Direct optimization is performed for \(\tau=\mathbf{u}\) from the Ego Planner, producing the optimized path \(\tau^*\). Only the safety cost is considered in this operation.

\textbf{Motion planning}:
This addresses per-step future states of the ego vehicle. Optimization is conducted using model predictive control (MPC) for control actions \(\mathbf{u}=[a,\delta]_{1:T}\) based on inverse dynamics: \(\mathbf{u}_t=\mathcal{T}^{-1}(\tau_{t+1},\tau_{t})\) \cite{hanselmann2022king}. The optimal motion planning is then transformed back through forward dynamics: \(\tau^*_{t+1}=\mathcal{T}(\tau^*_{t},\mathbf{u}^*_{t})\) after optimizations.

To solve this non-linear optimization problem, as illustrated in Equ. \ref{equ15}, we utilize the Gauss-Newton method \cite{bhardwaj2020differentiable}. that iteratively refines the planning variable as the initial value. The cost weights can be meticulously designed or learned directly, as the entire optimization process is fully differentiable \cite{huang2022differentiable}.

\section{Experiments}
\label{sec4}
In this section, we first introduce the experimental settings for the proposed HPP, including testing benchmarks, evaluation metrics, and detailed implementations. Subsequently, HPP is quantitatively compared against existing state-of-the-art methods and systems in predictions and planning. Discussions on ablation studies unveil the effectiveness and mechanism of modular co-design optimizations for HPP. Qualitative results further ablate the characteristics of HPP against certain state-of-the-art baselines.

\subsection{Experimental Setup}
\subsubsection{Testing Benchmarks}
To discover the comprehensive performances in predictions and planning for HPP, we summarize three questions to be tackled by benchmark testing: (1) How is the capabilities of HPP as full-stack ADS under interactive real-world cases? (2) How is the long-term horizon performance of HPP in interactive real-world scenarios? (3) How is the long-term driving functioning by HPP in continuous realistic scenarios? These prompt benchmarks accordingly:

(1) \textbf{nuScenes dataset} \cite{caesar2020nuscenes}\textbf{:} This dataset is among the largest and most widely used for full-stack autonomous driving. It includes over 1,000 20-second frames of driving scenarios annotated at 2 Hz, covering four cities worldwide. Benchmarked evaluations \cite{hu2023planning} involve 6,019 frames for all tasks of ADS under open-loop settings. Testing horizons are defined at $T=3\,\mathrm{s}$ and $T=6\,\mathrm{s}$ for motion predictions.

(2) \textbf{Waymo open motion dataset (WOMD)} \cite{ettinger2021large}\textbf{:} Utilized for long-term motion evaluations, it is the largest real-world dataset for interactive scenarios. It comprises 104,000 20-second frames representing unique scenarios, marked at 10 Hz, and encompasses over 570 km of driving and 1750 km of roadways. To assess long-term real-world performance in planning and motion predictions, evaluations are systematically conducted in the SMARTS benchmark \cite{huang2023gameformer}. This includes 400 highly interactive scenarios, each lasting 9 seconds, featuring representative behaviors. Autonomous vehicles are tasked with 5-second long-term planning and predictions in both open-loop and closed-loop configurations. Closed-loop testing involves leveraging the log simulator to replay driving scenarios for online interactions. To further examine the performance of joint predictions in HPP, the system is tested on the Waymo Occupancy Predictions benchmark \cite{mahjourian2022occupancy}. This involves predicting over 44,000 driving scenes of occupancy and flow within an 8-second timeframe at a frequency of 1 Hz.

\begin{table*}[htp]
\caption{nuScenes Open-loop Planning Testing Results ($^\dagger:$ Lidar inputs; $^\ddagger:$ Ego status; $^*:$ Augmentations)}
\centering
\small
\setlength{\tabcolsep}{5mm}{
\begin{tabular}{l|llll|llll}
\toprule
\multirow{2}{*}{Methods} &
  \multicolumn{4}{c|}{Collision rate (\%)$\downarrow$} &
  \multicolumn{4}{c}{Planning error (m)$\downarrow$} \\
               & @1 s    & @2 s    & @3 s      & Avg.   & @1 s    & @2 s    & @3 s   & Avg.    \\\midrule
NMP$^\dagger$ \cite{zeng2019end}   & -      & -      & 1.92  & -   & -      & -      & 2.31     & -  \\
SA-NMP$^\dagger$ \cite{zeng2019end}& -      & -      & 1.59  & -   & -      & -      & 2.05     & -       \\
FusionAD$^{\dagger\ddagger}$ \cite{ye2023fusionad}& \textbf{0.02}   & 0.08   & 0.27  & 0.12  & -    & -   & -     & 0.81   \\
FF$^{\dagger}$ \cite{hu2021safe}   & 0.06   & 0.17   & 1.07  & 0.43 & 0.55   & 1.20   & 2.54     & 1.43  \\
EO$^{\dagger}$ \cite{khurana2022differentiable} & 0.04   & 0.09   & 0.88  & 0.33 & 0.67   & 1.36   & 2.78     & 1.60   \\
OccNet \cite{tong2023scene}  & 0.21   & 0.59   & 1.37  & 0.72   & 1.29   & 2.13   & 2.99     & 2.13   \\
UniAD \cite{hu2023planning}  & 0.05   & \textbf{0.17}   & 0.71  & 0.31 & \textbf{0.48}   & 0.96   & 1.65     & 1.03  \\\midrule 
\rowcolor{mgray}
HPP$^\ddagger$  & 0.03  & \textbf{0.07} & 0.35 
& 0.15  & 0.30  & 0.61 & 1.15 & \textbf{0.72}  \\
\rowcolor{mgray}
\textbf{HPP}  & \textbf{0.03}  & \textbf{0.17} & \textbf{0.68} 
& \textbf{0.29} & \textbf{0.48}  & \textbf{0.91} & \textbf{1.54} & \textbf{0.97}  \\\midrule
Avg. Metrics       & @1 s    & @2 s    & @3 s      & Avg.   & @1 s    & @2 s    & @3 s   & Avg.    \\\midrule
ST-P3 \cite{hu2022st}   & 0.23   & 0.62   & 1.27  & 0.71 & 1.33   & 2.11   & 2.90     & 2.11   \\
VAD-Base$^{\ddagger}$  \cite{jiang2023vad}     & 0.07   & 0.10   & 0.24  & 0.14 & \textbf{0.17}   & \textbf{0.34}   & 0.60     & 0.37   \\
VAD-Base \cite{jiang2023vad} & 0.07   & 0.17   & 0.41  & 0.22 & \textbf{0.41}   & 0.70   & 1.05     & 0.72    \\
DeepEM$^*$ \cite{chen2023deepemplanner}&0.05   & 0.15   & 0.36  & 0.19 & 0.25   & 0.45  &  0.73     & 0.48   \\\midrule
\rowcolor{mgray}
HPP$^\ddagger$   & \textbf{0.02}  & \textbf{0.04} & \textbf{0.11}
& \textbf{0.06} & 0.26  & 0.37 & \textbf{0.59} & 0.40 \\
\rowcolor{mgray}
\textbf{HPP}  & \textbf{0.03}  & \textbf{0.08} & \textbf{0.24} 
& \textbf{0.12} & \textbf{0.41}  & \textbf{0.61} & \textbf{0.86} & \textbf{0.63}  \\\bottomrule
\end{tabular}
}
\label{table1}
\vspace{-0.1cm}
\end{table*}

(3) \textbf{CARLA simulator} \cite{dosovitskiy2017carla}\textbf{:} We utilize the Longest6 benchmark \cite{chitta2022transfuser} for long-term driving evaluations. The autonomous vehicle is assigned a $T=2\,\mathrm{s}$ horizons closed-loop planning task across 36 routes, ranging from 1.6 to 1.8 km, under various driving conditions in six CARLA towns.

\subsubsection{Testing Metrics} We adhere to the original benchmarks in the testing metrics configurations. Coinciding with the contributions of HPP, the devised metrics primarily focus on three aspects, i.e., Accuracy, Consistency, and Safety for predictions and planning. Detailed metrics are listed as follows:

(1) \textbf{Occupancy prediction:} Intersections over Union (IoU) and Area Under the Curve (AUC) \cite{chen2017deeplab} quantify the overall and per-grid prediction accuracy for occupancy. Video Panoptic Quality (VPQ) \cite{kim2020video} is adopted to assess the consistency of occupancy across marginal agents and perceptions.

(2) \textbf{Motion prediction:} Prediction accuracy is assessed using minimum average and final displacement errors (minADE, minFDE) for trajectories, as well as miss rate (MR) for each agent\cite{ettinger2021large}. Consistency is tested through joint displacement errors (JADE, JFDE) for all agents\cite{huang2023gameformer}, along with End-to-End Prediction Accuracy (EPA) \cite{gu2023vip3d} over perceptions.

(3) \textbf{Planning:} For open-loop testing, planning accuracy is evaluated using displacement errors (DE)\cite{hu2023planning} and the average distance \cite{jiang2023vad}. Consistency in predictions and safety is assessed through collision rates (CR) \cite{scheel2022urban}. In closed-loop settings, external measurements include infractions (IS), vehicle collisions (CV), routes completion (RC), and a driving score (DS) that encompasses overall driving performance \cite{dosovitskiy2017carla}.

\subsubsection{Implementation Details}
For fair comparisons, HPP is configured according to each benchmark with carefully devised learning pipelines and system architectures. In the nuScenes dataset, full-size training is conducted with a total batch size of 4. For planning, 10\% of the full training set is randomly sampled, and the full set is utilized for the occupancy benchmark in WOMD. Learning occurs in batches of 24. The expert dataset from the CARLA benchmark \cite{renz2022plant}, collected in different towns, is directly used for training in batches of 32.

Training strategies for all benchmarks are aligned using a distributed strategy on four NVIDIA A100 GPUs. The AdamW optimizer is employed with an initial learning rate of 1e-4, and a cosine annealing learning rate strategy is applied. The total number of training epochs is set to 20. We apply the same GPU devices for nuScenes and WOMD for testing. Evaluations for the CARLA benchmark are conducted in one NVIDIA RTX 3080 GPU.

For system architectures, HPP establishes BEV perception ego-centered within $\pm$50 m of $H, W=200$ in nuScenes. In WOMD and CARLA benchmarks, privileged perceptions are assumed. Therefore, HPP is developed by encoding perfect scene context inputs according to our previous work \cite{huang2023gameformer, liu2021multimodal}. For CARLA benchmarks, BEV remains ego-centered within $\pm32$ m of $H, W=128$. BEV settings for WOMD follow the official benchmark guidelines. Path planning is optimized without knowledge of reference routes in nuScenes and CARLA. In WOMD, motion planning is conducted considering reference information. The full-stack HPP in nuScenes considers various queries for agents. In WOMD and CARLA, agents are sorted and filtered to $N_A=11$. We select the ReLU activation function and apply a dropout rate of 0.1. We refer more detailed parameters with notations in Table \ref{table12}. 

\subsection{Main Results}
\subsubsection{Full-stack ADS Performance} We report HPP's testing performance against recent state-of-the-art methods on the nuScenes dataset. HPP has achieved \emph{state-of-the-art} results across various key metrics in both prediction and planning.

\begin{table*}[tp]
\caption{Open-loop Planning Compared with LLMs ($^\ddagger:$ Ego status)}
\centering
\small
\setlength{\tabcolsep}{5mm}{
\begin{tabular}{l|llll|llll}
\toprule
\multirow{2}{*}{Methods} &
  \multicolumn{4}{c|}{Collision rate (\%)$\downarrow$} &
  \multicolumn{4}{c}{Planning error (m)$\downarrow$} \\
               & @1 s    & @2 s    & @3 s      & Avg.   & @1 s    & @2 s    & @3 s   & Avg.    \\\midrule
GPTDriver$^\ddagger$  \cite{mao2023gpt}   & 0.07   & 0.15   & 1.10  & 0.44  & 0.27   & 0.74   & 1.52     & 0.84   \\
Agent-Driver$^\ddagger$ \cite{mao2023language}      & \textbf{0.02}   & 0.13   & 0.48  & 0.21 & \textbf{0.22}   & 0.65   & 1.34     & 0.74 \\\midrule
\rowcolor{mgray}
\textbf{HPP}$^\ddagger$    & 0.03  & \textbf{0.07} & \textbf{0.35} & 0.30  & \textbf{0.61} & \textbf{1.15} & \textbf{0.72} 
& \textbf{0.15} \\\bottomrule
\end{tabular}
}
\label{table2}
\vspace{-0.2cm}
\end{table*}

\begin{table}[tp]
\centering
\caption{Testing Results on nuScenes Occupancy Prediction}
\begin{tabular}{l|>{\columncolor[gray]{0.9}}llll}
\toprule
Methods                               & \textbf{IoU-n.} $\uparrow$           & IoU-f. $\uparrow$            & VPQ-n. $\uparrow$     & VPQ-f. $\uparrow$ \\\midrule
FIERY\cite{hu2021fiery}       & 59.4     & 36.7    & 50.2   &  29.9    \\
StretchBEV\cite{akan2022stretchbev} & 55.5               & 37.1         & 46.0                & 29.0    \\
ST-P3\cite{hu2022st} & -               & 38.9         & -                & 32.1   \\
BEVerse\cite{zhang2022beverse} & 61.4                & \textbf{40.9}         & 54.3                & \textbf{36.1}    \\
PowerBEV\cite{li2023powerbev} & 62.5                & 39.4         & 55.5                & 33.8    \\
UniAD\cite{hu2023planning} & 63.4                & 40.2         & 54.7                & 33.5    \\
\midrule
\textbf{HPP}                        & \textbf{64.8}                & 40.5         & \textbf{56.4}            & 34.7   \\\bottomrule
\end{tabular}
\label{table3}
\end{table}

(1) \textbf{Planning results:} Described in Table \ref{table1}, HPP achieves state-of-the-art (SOTA) results across all planning horizons (@1 s-@3 s) against various autonomous driving systems in both absolute metrics \cite{hu2023planning} and average ones \cite{jiang2023vad}. Specifically, HPP presents a $\textbf{5.9\%}$ lower average L2 errors and a $\textbf{6.5\%}$ lower average collision rate compared to UniAD \cite{hu2023planning}, utilizing identical BEV perceptions. This showcases the validity of the modular co-design for HPP in predictions and planning.

In average metrics, HPP reports a $\textbf{12.5\%}$ improvement in planning errors against VAD \cite{jiang2023vad} which highlights superior perception modules. The reasoning design for HPP manifests with a $\textbf{30\%}$ lower collision rate compared to DeepEM \cite{chen2023deepemplanner}, which also features reasoning by EM decoding and extra de-noising augmentations. Compared with methods leveraging ego-status $\mathbf{s}_{\text{ego}}$, a variant of HPP adding ego-status (denoted HPP$^\ddagger$) in Ego Planner (Sec. \ref{sub5}) is trained with boasting results. HPP$^\ddagger$ does not include accelerations in \cite{jiang2023vad} as leakage of ground-truths. In comparison with FusionAD \cite{ye2023fusionad}, which depends on excessive LIDAR fusion, HPP also demonstrates an $\textbf{11.8\%}$ lower planning error with comparable safety. Our system also exhibits $\textbf{2.8\%}$ lower errors and nearly $\textbf{20\%}$ lower collision rates compared to LLM methods (see Table \ref{table2}). This further substantiates the effectiveness of modular integration of predictions in planning-oriented objectives, as LLM baselines focus more on alignments by language knowledge. 

(2) \textbf{Predictions results:} Joint results of occupancy predictions are tested in two ranges (near: $30\times30$ m; far: $50\times50$ m) centered on the autonomous vehicle. Shown in Table \ref{table3}, HPP presents advanced accuracy $\textbf{+2.5\%}$ and consistency $\textbf{+3.7\%}$ compared to \cite{hu2023planning}, thanks to proposed Ms-OccFormer that integrate mutually with motion predictions. Validity of modular co-design in HPP is further manifested by $\textbf{+4\%}$ and $\textbf{+1.6\%}$ improved IoU and VPQ without extra augmentations, compared with \cite{li2023powerbev} learning single occupancy task. 

Marginal results of motion predictions are presented in Table \ref{table4}. Here, we compare the prediction results averaged from all vehicles (-v.) and measure full agent (-f.) results weighted by categories. HPP reports a \textbf{2.8\%} lower minADE and a \textbf{+3.2\%} EPA gain in vehicle predictions, with \textbf{3.7\%} and \textbf{+6\%} improved performance predicting all agents compared to baselines \cite{hu2023planning} under the same perception settings. This highlights the performance gain achieved through game-theoretic reasoning and joint dependencies in GTFormer (Sec. \ref{sub4}).

\begin{table}[tp]
\centering
\caption{Testing Results on nuScenes Motion Predictions}
\begin{tabular}{l|>{\columncolor[gray]{0.9}}llll}
\toprule
Methods                               & \textbf{minADE} $\downarrow$           & minFDE $\downarrow$            & MR. $\downarrow$     & EPA $\uparrow$ \\\midrule
ViP3D\cite{gu2023vip3d}       & 1.15     & 1.95    & 22.6   &  0.222    \\
PnPNet\cite{liang2020pnpnet} & 2.05              & 2.84         & 24.6                &  0.226   \\
UniAD-v.\cite{hu2023planning} & 0.71                & 1.02        & 15.1               & 0.456    \\
UniAD-f.\cite{hu2023planning} & 0.911                & 1.236        & 15.1               & 0.314    \\
\midrule
\textbf{HPP-v.}                        & \textbf{0.682}                & \textbf{0.947}         & \textbf{13.8}           & \textbf{0.471}   \\
\textbf{HPP-f.}                        & 0.878                & 1.205         & 14.5            & 0.334   \\\bottomrule
\end{tabular}

\label{table4}
\end{table}

\begin{table}[tp]
\centering

\caption{Testing Results on Waymo Occupancy Flow Benchmark}
\setlength{\tabcolsep}{3mm}{
\begin{tabular}{l|>{\columncolor[gray]{0.9}}llll}
\toprule
\multirow{2}{*}{Methods}   & \textbf{AUC}   & AUC  & EPE & AUC \\
& \textbf{-obs.} $\uparrow$ & -occ. $\uparrow$ & -f. $\downarrow$ & -FT $\uparrow$\\
\midrule
MotionPerceiver                      & 77.1    & -  & -   &  -  \\
OFMPNet                              & 77.0    & 16.5   &  3.58&  76.1\\
STCNN                                & 74.4    & 16.8   &  3.87&  73.3\\
HOPE\cite{hu2022hope}           & \textbf{80.3}  & 16.5   &  3.67 & \textbf{83.9}\\
STrajNet\cite{liu2023multi}       & 77.8     & 17.8   &  3.20  & 78.5\\
VectorFlow\cite{huang2022vectorflow} & 75.4     & 17.3    & 3.58 & 76.7\\
\midrule
\textbf{HPP}                         & 79.7     & \textbf{19.4}  &\textbf{2.95} & 80.2\\\bottomrule
\end{tabular}
}
\label{table5}
% \vspace{-0.1cm}
\end{table}

It is important to note that HPP focuses on predictions and planning rather than perception. Prediction results could potentially be further enhanced with better perceptions \cite{jiang2023vad} or additional LIDAR inputs \cite{ye2023fusionad}. However, HPP still outperforms in final planning results, demonstrating better social compliance denoted as our contributions.

\subsubsection{Long-horizon Interactive Performance} \label{womd_test} We benchmark HPP's long-horizon performance under WOMD. HPP has underscored advanced performances against numerous SOTA methods in predictions and planning. 

(1) \textbf{Occupancy results:} Shown in Table \ref{table5}, HPP demonstrates superior prediction accuracy (\textbf{+8.8\%} AUC-occ.) and lower flow error (\textbf{9.1\%} EPE-f.) when considering joint flow predictions. The explicit design of MS-OccFormer in HPP has proven its strong accuracy compared to our previous work \cite{liu2021multimodal}, which only conducts global interactions without marginal awareness. HPP presents a \textbf{+2.4\%} improvement in AUC-obs. and \textbf{+2.2\%} in flow-traced occupancy.

\begin{table*}[tp]
\caption{Open-loop Planning Testing Results on WOMD Scenarios (SMARTS Benchmark)}
\centering
\small
\setlength{\tabcolsep}{3.5mm}{
\begin{tabular}{l|l|l|lll|ll}
\toprule
\multirow{2}{*}{Method} &
  \multicolumn{1}{c|}{Collision rate}  &
  \multicolumn{1}{c|}{Miss rate}  &
  \multicolumn{3}{c|}{Planning error (m) $\downarrow$} &
  \multicolumn{2}{c}{Prediction error (m) $\downarrow$} \\
               & \multicolumn{1}{c|}{(\%) $\downarrow$}&\multicolumn{1}{c|}{(\%) $\downarrow$}  & @1 s            & @3 s            & @5 s           & JADE           & JFDE \\ \midrule
Vanilla IL \cite{huang2022differentiable}     & 4.25          & 15.61            & 0.216          & 1.273          & 3.175          & --            & --       \\
DIM  \cite{rhinehart2018deep}          & 4.96          & 17.68            & 0.483          & 1.869          & 3.683          & --            & --         \\
OPGP  \cite{liu2023occupancy}          & 3.79         & 12.89            & 0.245          & 1.672          & 3.099          & --            & --         \\
MultiPath++ \cite{varadarajan2022multipath++}   & 2.86          & 8.61             & 0.146          & 0.948          & 2.719          & --            & --       \\
MTR-e2e  \cite{shi2022motion}      & 2.32          & 8.88             & 0.141          & 0.888          & 2.698          & --            & --     \\
DIPP  \cite{huang2022differentiable}         & 2.33          & 8.44             & 0.135          & 0.928          & 2.803          & 0.925         & 2.059       \\
GameFormer  \cite{huang2023gameformer}         & 1.98 & \textbf{7.53}    & 0.129 &\textbf{0.836}  & \textbf{2.451} &0.853 & \textbf{1.919} \\ \midrule 
\rowcolor{mgray} \textbf{HPP}           & \textbf{1.85} & 7.58    & \textbf{0.092} & 0.881  & 2.667 &\textbf{0.829} & 1.965 \\ \bottomrule
\end{tabular}
}
\vspace{-0.3cm}
\label{table6}
\end{table*}

\begin{table*}[htp]
\caption{Close-loop Planning Testing Results on WOMD Scenarios (SMARTS Benchmark)}
\centering
\resizebox{\linewidth}{!}{
\begin{tabular}{l|ll|lll|lll}
\toprule
\multirow{2}{*}{Method} &
\multicolumn{1}{c}{Success rate} &
\multicolumn{1}{c|}{Progress} &
\multicolumn{1}{c}{Acceleration} &
\multicolumn{1}{c}{Jerk} &
\multicolumn{1}{c|}{Lateral acc.} &
\multicolumn{3}{c}{Position error to expert driver ($\mathrm{m}$) $\downarrow$} \\
                    &  \multicolumn{1}{c}{(\%) $\uparrow$}  & \multicolumn{1}{c|}{$(\mathrm{m})$ $\uparrow$ }  & \multicolumn{1}{c}{($\mathrm{m}/\mathrm{s}^2$) } & \multicolumn{1}{c}{($\mathrm{m}/\mathrm{s}^3$)} & \multicolumn{1}{c|}{($\mathrm{m}/\mathrm{s}^2$)} & @3 s                   & @5 s            & @8 s \\ \midrule
\emph{Expert} & - & 54.52 & 0.529 & 1.020 & 0.103 & - & - \\
Vanilla IL \cite{huang2022differentiable}         & 0                     & 6.23                  & 1.588                 & 16.24                 & 0.661                 & 9.355                 & 20.52          & 46.33  \\
RIP \cite{rhinehart2018deep}             & 19.5                  & 12.85                 & 1.445                 & 14.97                 & 0.355                 & 7.035                 & 17.13          & 38.25  \\
CQL \cite{kumar2020conservative}   & 10                    & 8.28                  & 3.158                 & 25.31                 & 0.152                 & 10.86                 & 21.18          & 40.17   \\\midrule
DIPP \cite{huang2022differentiable}            & 68.12$\pm$5.51        & 41.08$\pm$5.88        & 1.44$\pm$0.18         & 12.58$\pm$3.23        & \textbf{0.31}$\pm$0.11         & 6.22$\pm$0.52         & 15.55$\pm$1.12 & 26.10$\pm$3.88 \\
GameFormer  \cite{huang2023gameformer}   & 73.16$\pm$6.14        & 44.94$\pm$7.69        & \textbf{1.19}$\pm$0.15         & 13.63$\pm$2.88        & 0.32$\pm$0.09         & 5.89$\pm$0.78         & 12.43$\pm$0.51 & 21.02$\pm$2.48 \\
\rowcolor{mgray}
\textbf{HPP}                & \textbf{74.28}$\pm$5.49        & \textbf{47.17}$\pm$8.92        & 1.33$\pm$0.18         & \textbf{11.68}$\pm$2.76        & 0.35$\pm$0.09         & \textbf{4.93}$\pm$0.78         & \textbf{10.24}$\pm$0.82 & \textbf{18.99}$\pm$3.05 \\\midrule
DIPP (optim.)& 92.16$\pm$0.62        &51.85$\pm$0.14         & 0.58$\pm$0.03         & \textbf{1.54}$\pm$0.19& 0.11$\pm$0.01         & 2.26$\pm$0.10         & 5.55$\pm$0.24  & 12.53$\pm$0.48 \\
GameFormer (optim.)&\textbf{94.50}$\pm$0.66&\textbf{52.67}$\pm$0.33& \textbf{0.53}$\pm$0.02& 1.56$\pm$0.23         & \textbf{0.10}$\pm$0.01&\textbf{2.11}$\pm$0.21 & \textbf{4.87}$\pm$0.18 &\textbf{11.13}$\pm$0.33  \\
\rowcolor{mgray}
\textbf{HPP (optim.)}        & 92.25$\pm$0.85        & 52.19$\pm$0.41        & 0.66$\pm$0.02         & 1.87$\pm$0.28        & 0.10$\pm$0.01         & 2.13$\pm$0.29         & 4.90$\pm$0.26 & 12.89$\pm$0.38 \\\bottomrule
\end{tabular}
}
\vspace{-0.3cm}
\label{table7}
\end{table*}

(2) \textbf{Open-loop results:} Table \ref{table6} reveals the compelling results of HPP over numerous state-of-the-art (SOTA) methods in open-loop prediction and planning. HPP achieves notable reductions of over \textbf{50\%} in collisions and \textbf{26\%} in planning errors compared to imitation learning baselines \cite{huang2022differentiable, rhinehart2018deep} that discount predictions. When compared with SOTA motion prediction baselines as imitative planners, HPP also demonstrates \textbf{2.3\%} to \textbf{17.6\%} reductions in planning errors and \textbf{21.1\%} fewer collisions. This underscores the importance of integrated predictions and planning.

\begin{figure}[tp]
    \centering
    \includegraphics[width=\linewidth]{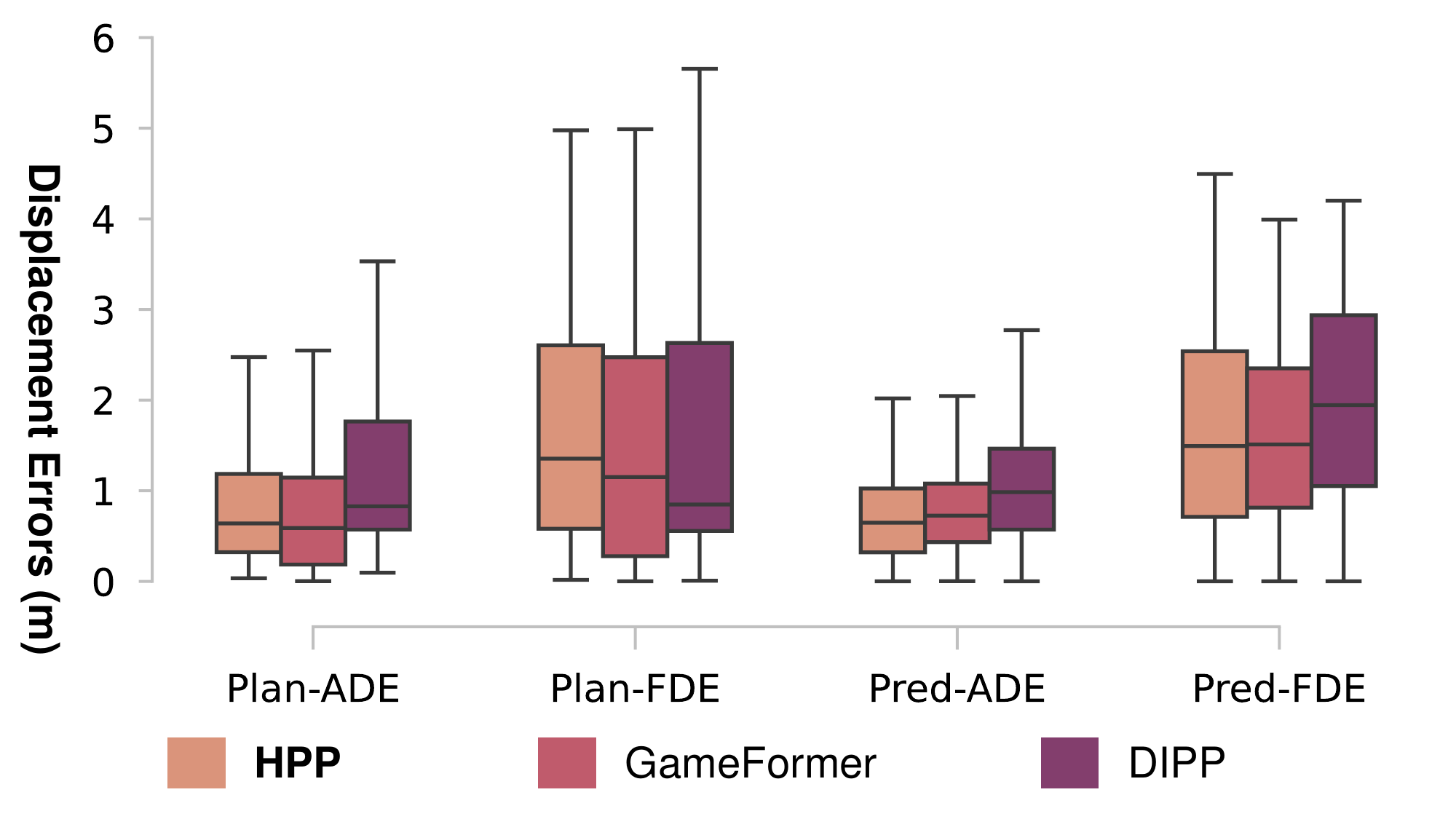}
    \caption{Open-loop predictions and planning results for IPP baselines in WOMD. HPP presents the best short-term performance with lower variance.} 
    \label{fig5}
    % \vspace{-0.2cm}
\end{figure}

In comparison with integrated baselines, HPP presents significant improvements over the IOP framework \cite{liu2023occupancy} due to the absence of future interactions for planning and the intractable occupancy in IOP that hampers normal guidance. When compared with IPP systems, HPP delivers superior performance in collision rates (\textbf{6.6\%}), joint prediction errors (\textbf{2.9\%}), and comparable planning errors. Compared to the state-of-the-art results from our previous works \cite{huang2022differentiable,huang2023gameformer}, which focus on modeling future interactions and reasoning (see Fig. \ref{fig5}), HPP exhibits superior short-term performance with reduced variance. This further validates the efficacy of co-design integration for joint predictions and planning upon reasoning in HPP.

(3) \textbf{Closed-loop results:} In Table \ref{table7}, HPP undergoes testing in a replay simulator against state-of-the-art IPP systems \cite{huang2023gameformer}, IL methods \cite{rhinehart2018deep}, and RL approaches \cite{kumar2020conservative}. IL and RL results are significantly compromised due to accumulated distributional shifts in closed-loop interactions with the environment. HPP exhibits a higher success rate (\textbf{+1.5\%}) and lower planning errors (\textbf{13.2\%}) compared to our previous methods \cite{huang2023gameformer}, thanks to the hybrid prediction-aware planner designed conditioned on goal context. The closed-loop performance sees substantial improvements with online optimizations. Due to the high cost of the occupancy process in raw data, HPP is re-planned by online refinements at 2 Hz. Testing results demonstrate strong performance against our previous methods \cite{huang2023gameformer}, which re-plan more frequently at 10 Hz, achieving \textbf{8.5\%} lower closed-loop positional errors by \cite{huang2022differentiable}.

\begin{table}[tp]
\centering
\caption{Testing Results on CARLA Longest6 Benchmark}
\setlength{\tabcolsep}{4mm}{
\begin{tabular}{l|>{\columncolor[gray]{0.9}}llll}
\toprule
Methods                               & \textbf{DS} $\uparrow$  & RC$^*$ $\uparrow$  & IS $\uparrow$  & CV $\downarrow$ \\\midrule
Rule-Based\cite{dosovitskiy2017carla}      & 38.0                    & 29.1           & 0.84          & 0.64    \\
KING\cite{hanselmann2022king}        & 45.1                    & 78.3           & 0.55          & 1.67    \\
Roach\cite{zhang2021end}       & 55.3                    & \textbf{88.2}           & 0.62          & 0.72    \\
PlanT$^*$\cite{renz2022plant}       & \textbf{70.9}                    & 83.1           & \textbf{0.87}          & \textbf{0.31}    \\
% \emph{Expert}$^*$                         & 66.4                    & 78.8           & 0.86          & 0.28    \\
\midrule
\textbf{HPP$^*$}                        & 65.5              & 79.2         & 0.82           & 0.51   \\\bottomrule
\end{tabular}
}
\label{table8}
\end{table}

\begin{table*}[t]
\caption{Ablation study on modular co-design integration among predictions and planning}
\centering
\resizebox{\linewidth}{!}{
\begin{tabular}{l|lll|lll|ll|>{\columncolor[gray]{0.9}}llll|>{\columncolor[gray]{0.9}}lll|ll}
\toprule
\multirow{2}{*}{ID} &\multicolumn{3}{c|}{MS-OccFormer(wo.)} & \multicolumn{3}{c|}{GTFormer(wo.)}  & \multicolumn{2}{c|}{EgoPlanner (wo.)}  &
  \multicolumn{4}{c|}{Occupancy Prediction} &
  \multicolumn{3}{c|}{Motion Prediction} & \multicolumn{2}{c}{Planning} \\
  & Score& Motion& AC-Occ.& Score& Motion& Occ.& Motion & Occ.&IoU-n.$\uparrow$ & IoU-f.$\uparrow$ &VPQ-n.$\uparrow$   & VPQ-f.$\uparrow$ & minADE$\downarrow$ & MR$\downarrow$    & EPA$\uparrow$   & ADE$\downarrow$ & CR$\downarrow$\\\midrule
1& \XSolidBrush  & - & - & - & - & - & - & - & 63.5 & \textbf{40.0} & 54.5 & \textbf{33.7} & 0.733 & 16.1 & 45.6 & 0.989 & 0.428 \\
2& - & \XSolidBrush  & - & - & - & - & - & - & 63.1 & 39.4 & 53.6 & 32.8 & 0.739 & 16.2 & 44.8 & 0.994 & 0.441 \\
3& -  & \XSolidBrush  & \XSolidBrush & - & - & - & - & - & 61.9 & 39.6 & 50.8 & 30.9& 0.748 & 16.6 & 43.8 & 1.015 & 0.466 \\\midrule
4& - & - & - & \XSolidBrush  & - & - & - & - & 63.7 & 39.8 & 55.0 & 33.3 & 0.728 & 15.9 & 45.0 & 0.986 & 0.402 \\
5& - & - & - & - & \XSolidBrush & - & - & - & 63.5 & 39.5 & 54.6 & 33.5 & 0.734 & 16.4 & 44.8 & 0.992 & 0.432 \\
6& - & - & - & - & \XSolidBrush & \XSolidBrush  & - & -  & 63.5 & 39.6 & 54.8 & 33.0 & 0.747 & 16.2 & 44.2 & 0.998 & 0.436 \\\midrule
7& - & - & - & - & - & - & \XSolidBrush & - & 63.6 & 39.8 & 55.0 & 33.4 & 0.726 & 16.0 & 45.7 & 0.989 & 0.414 \\
8& - & - & - & - & - & - & - & \XSolidBrush & 63.5 & 39.7 & 54.7 & 33.3 & \textbf{0.721} & \textbf{15.6} & 45.4 & 1.012 & 0.458 \\
9& - & - & - & - & - & - & \XSolidBrush & \XSolidBrush & 63.5 & 39.8 & 54.8 & 33.4 & 0.726 & 15.9 & 45.7 & 0.997 & 0.449 \\\midrule
0& - & - & - & - & - & - & -& - & \textbf{63.7} & 39.8 & \textbf{55.0} & 33.5 & 0.724 & 15.9 & 45.8 & \textbf{0.985} & \textbf{0.402} \\\bottomrule
\end{tabular}
\label{table9}
\vspace{-0.2cm}
}
\end{table*}
\subsubsection{Long-term Driving Performance} HPP demonstrates comparable driving capabilities against the state-of-the-art method \cite{renz2022plant} in the CARLA benchmark (Table \ref{table8}). With significant improvements over rule-based agents \cite{dosovitskiy2017carla}, HPP achieves a $+20.4$ driving score compared to IPP methods \cite{hanselmann2022king}, which guide planning by adversarial predictions, as well as a $+12.2$ improvement compared to RL methods \cite{zhang2021end}. Noted that both HPP and the reproduced \cite{renz2022plant} ($^*$) show compromised route completions (RC), likely due to GPU inference issues\footnote{\url{https://github.com/autonomousvision/plant/issues/17}}.

\begin{figure}[tp]
    \centering
    \includegraphics[width=\linewidth]{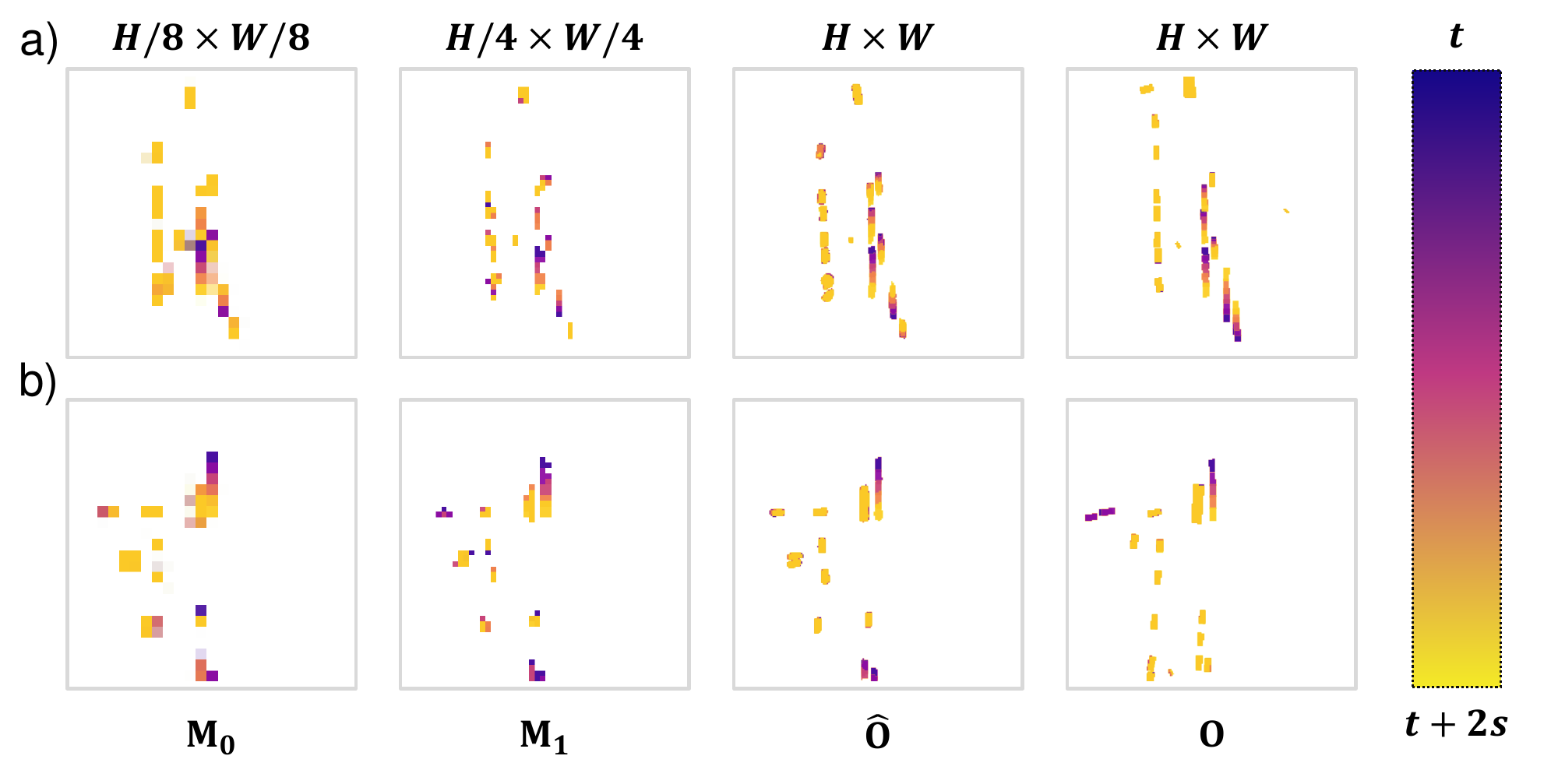}
    \caption{Qualitative ablations of multi-scale attention masks $\mathbf{M}_{1:T}$ compared with occupancy predictions $\hat{\mathbf{O}}_{1:T}$ and the ground-truth $\mathbf{O}_{1:T}$. Aligned results from multiple granularities under scenarios a) and b) reveal the validity of multi-scale prediction-wise integration design in MS-OccFormer (Sec. \ref{multiscale}).}  
    \label{fig6}
    % \vspace{-0.2cm}
\end{figure}

\subsection{Ablation Studies and Discussions}
To unleash the internal effectiveness in HPP, comprehensive ablation studies are conducted in nuScenes centered on discussing the roles in modular co-design and characteristics for hybrid prediction in planning guidance. 

\subsubsection{Roles in Modular Co-design} To elicit the effectiveness of co-design among each module, as presented in Table \ref{table9}, we purposefully remove certain key designs for integration across each sub-module. Referred to ID 1-9, all ablated baselines are compared to the original HPP (ID.0). Detailed discussions are as follows:

(1) \textbf{Effects in Ms-OccFormer:} For marginal dependencies (Sec. \ref{margin_dep}), when removing score scaling (ID.3 vs. ID.0), we observe a slight trade-off under occupancy range (-0.5 VPQ-n. vs. +0.2 VPQ-f.), with increased collisions ($+0.026\%$). This indicates the scaling captures multi-modal features for joint predictions and enhances the shared occupancy area nearby. Removing motion fusion (ID.2) causes inclusive decreases for lack of marginal alignments. Greater decreases are observed (-1.2 IoU-n. and -3.7 VPQ-n.) in removing agent-conditioned occupancy modeling. It highlights the importance of future interactions and validates the consistency towards motion predictions ($2.4\%$ lower for prediction errors).  Results in planning also reflect more contributions in near-scale joint prediction interactions to planning. Comparing ID.2 and ID.3, without sacrificing far-scale occupancy (-1.2 IoU.-n vs. +0.2 IoU-f.), the planning errors and collisions increase along with a drop in near-scale prediction accuracy. For the key design of multi-scale predictions integration, qualitative (see Fig. \ref{fig6}) and quantitative ablations (Table \ref{table10}) have demonstrated notable improvements from multi-scale attention mask update design (+1.2 IoU-n.) as well as local integration (+1.0 VPQ-n.) for prediction consistency.  

(2) \textbf{Effects in GTFormer:} As the core module that enables reasoning capabilities for consistent predictions and planning, we try to discover roles by removing marginal and joint predictions in GTFormer. Compared ID.4 with ID.5, the removal of the reasoning module had a thorough drop, especially planning ($+0.03\%$ CR) and motion predictions ($+3.1\%$ MR). This highlights the importance of reasoning in consistency and accuracy for future interactions. Meanwhile, considering ID.5 and ID.6, the increase of prediction errors ($+1.7\%$ minADE) implies that joint dependencies (Sec. \ref{joint_dep}) are the key in consistent motion predictions upon GT-reasoning. This reflects an enhanced mutual consistency modulated by interactive modeling between joint and marginal predictor co-design.

\begin{table}[t]
\centering
\caption{Ablations on  multi-scale prediction-wise integration}
\begin{tabular}{l|>{\columncolor[gray]{0.9}}llll}
\toprule
Baselines     & \textbf{IoU-n.} $\uparrow$           & IoU-f. $\uparrow$            & VPQ-n. $\uparrow$     & VPQ-f. $\uparrow$ \\\midrule
w/o. global integration    & 61.6     & 38.8    & 52.0   &  31.9    \\
w/o. local integration  &    61.8       & 38.8         & 52.8    & 32.1   \\
w/o. attn. mask $\textbf{M}_t$  & 62.5     & 39.0         & 53.8       & 32.4    \\
\midrule
\textbf{HPP}                        & \textbf{63.7}                & 39.8         & \textbf{55.0}            & 33.5   \\\bottomrule
\end{tabular}
\label{table10}
\end{table}

\begin{figure*}[ht]
    \centering
    \includegraphics[width=\linewidth]{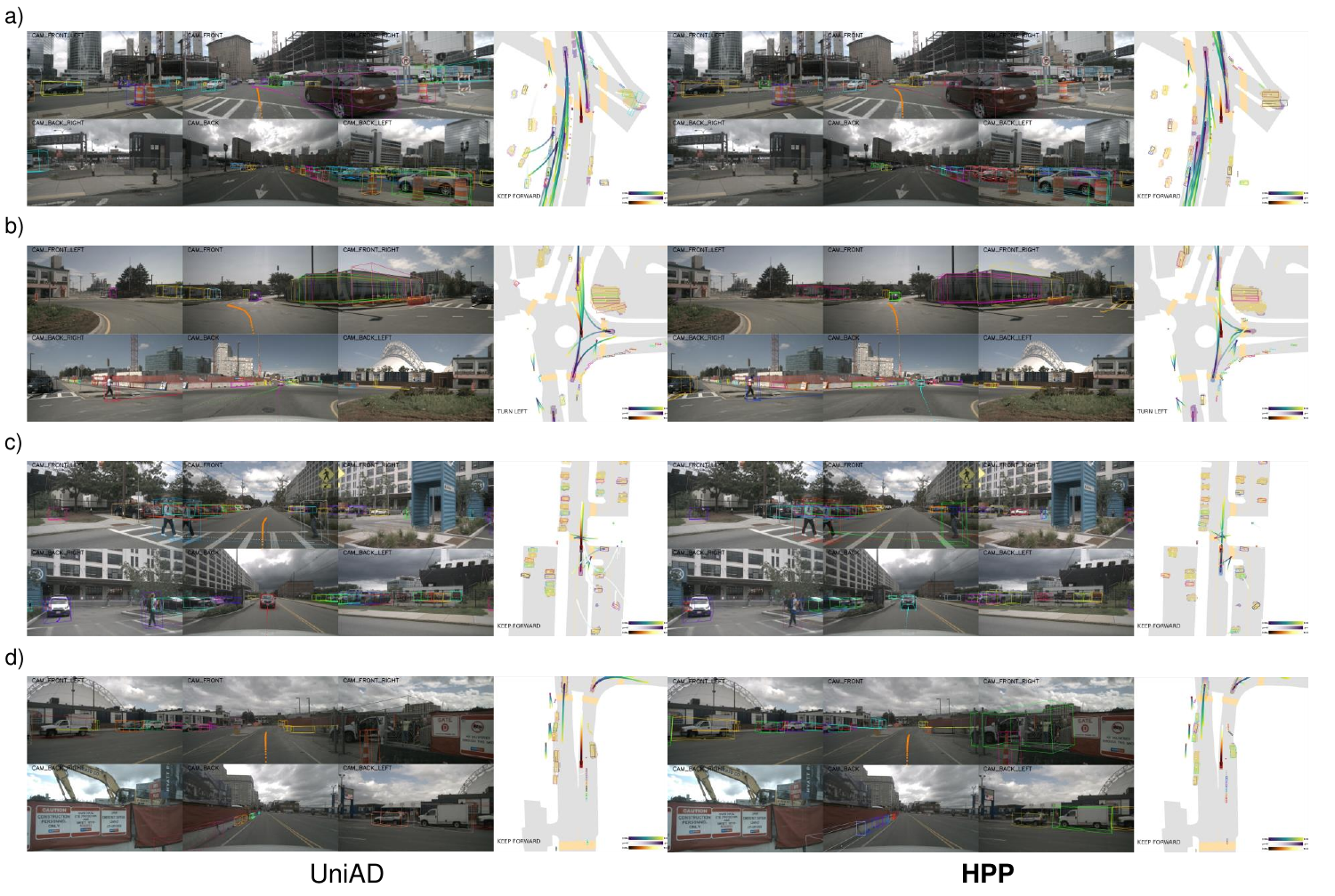}
    \caption{Qualitative results of full-stack open-loop testing on nuScenes \cite{caesar2020nuscenes}. Each Sub-figure renders hybrid prediction (in green and violet) and planning (in red) under camera view (left with rendered trajectories and bounding boxes) and BEV space (right). HPP is compared with \cite{hu2023planning} under several driving scenarios featuring specific interactions: a) Cruising through intersections; b) Driving across a crowded roundabout; c) Yielding to pedestrians on crosswalks; d) Lane keeping in fork roads.} 
    \label{fig7}
    % \vspace{-0.5cm}
\end{figure*}

\begin{table*}[htbp]
\centering
\caption{Ablation study on hybrid prediction guidance for planning}
\setlength{\tabcolsep}{3.6mm}{
\begin{tabular}{l|llll|lll>{\columncolor[gray]{0.9}}l|lll>{\columncolor[gray]{0.9}}l}
\toprule
\multirow{2}{*}{ID} &\multirow{2}{*}{$\mathcal{L}_{col}$}       & \multirow{2}{*}{Occ.}  & \multirow{2}{*}{Most-likely} &  \multirow{2}{*}{Full-motion}  &
  \multicolumn{4}{c|}{Collision rate (\%) $\downarrow$} &
  \multicolumn{4}{c}{Planning error (m) $\downarrow$} \\
  & & & & & @1 s    & @2 s    & @3 s      & Avg.   & @1 s    & @2 s    & @3 s   & Avg.  \\\midrule
1&-             & -     & -    & -    & 0.11  & 0.31  & 0.73  & 0.39  & 0.19  & 0.53   & 1.13  & 0.61     \\
2&\Checkmark    & -     & -    & -    & 0.07  & 0.22  & 0.64  & 0.31  & \textbf{0.18}  & \textbf{0.52}   & \textbf{1.09}  & \textbf{0.60}  \\ 
3&\Checkmark & - &\Checkmark & -& 0.05  & 0.11  & 0.60  & 0.25  & 0.19  & \textbf{0.52}  & 1.10 & \textbf{0.60}  \\
4&\Checkmark & -&-& \Checkmark & 0.04   & 0.09  & 0.53 & 0.20 & 0.19  & \textbf{0.52} & 1.10 & \textbf{0.60}  \\
5&\Checkmark &\Checkmark& -    & -      & 0.03  & 0.14  & 0.45  & 0.20  & 0.30  & 0.61   & 1.16  & 0.72   \\
6&\Checkmark & \Checkmark&\Checkmark & -& 0.03  & 0.14  & 0.44 & 0.20  & 0.31  & 0.61   & 1.17 & 0.72   \\\midrule
0&\Checkmark& \Checkmark&- &\Checkmark& \textbf{0.03}  & \textbf{0.07}   & \textbf{0.35}  & \textbf{0.15}     & 0.30  & 0.61   & 1.15  & 0.72 \\\bottomrule
\end{tabular}
\label{table11}
\vspace{-0.1cm}
}
\end{table*}

(3) \textbf{Effects in Ego Planner:} Validating the roles of plan-conditioning design compared to our previous work \cite{huang2023gameformer} in Sec. \ref{womd_test}, we further examine the effects of hybrid prediction interactions. In ID.9, we observe marginal improvements in predictions (+0.2 IoU, +0.1 EPA) and more significant enhancements ($-0.047\%$ CR) in planning with the inclusion of hybrid prediction interactions. This underscores the consistency modeling contributed along with the HPP design. Ablations from ID.7 and ID.8 further suggest a substantial impact of interactions with occupancy predictive features ($-0.056\%$ CR) over marginal ones ($-0.012\%$ CR) on planning. Solely motion conditional planning (ID.8) outputs overly optimistic motion predictions (-0.3 MR). This in turn harms the original game-theoretic reasoning, resulting in inferior planning ($+0.035\%$ CR). These results underscore the modulating effect of joint dependencies in both consistencies for planning and motion predictions. 

\subsubsection{Roles in hybrid prediction}
To further delve into the characteristics of marginal and joint predictions, listed in Table \ref{table11}, the original HPP (ID.0) is measured with ablations (ID.1-6) guiding: reasoning loss ($\mathcal{L}_{col}$); marginal predictions (most likely, full); and joint predictions (Occ.) during learning and optimizations. Key findings are summarized below:

(1) \textbf{Consistent reasoning:} Comparing ID.1 and ID.2, significant safety improvements ($-25.8\%$ CR) highlights the consistency role for marginal predictions in reasoning learning ($\mathcal{L}_{col}$). Joint predictions in HPP are leveraged to guide internal marginal consistency that modulates planning. 

(2) \textbf{Complementary influences:} hybrid prediction benefit planning in mutual coverage gains compared with sole guidance ($-33\%$ CR) in ID.4 and ID.5. This is enhanced upon consistency design in HPP. 1) Compared with ID.3, guiding with full motion predictions (ID.2) spawns safe planning ($-20\%$ CR) without sacrificing accuracy. 2) A good alignment in hybrid prediction results in close performance in ID.5 and ID.6 adding the most likely marginal prediction. These mutual effects ascertain the planning performance in case either of the hybrid prediction is less functioning due to: (1) state error for pose and speed; (2) future uncertainty; or (3) discontinuation under spatial temporal horizons.

\begin{figure*}[ht]
    \centering
    \includegraphics[width=\linewidth]{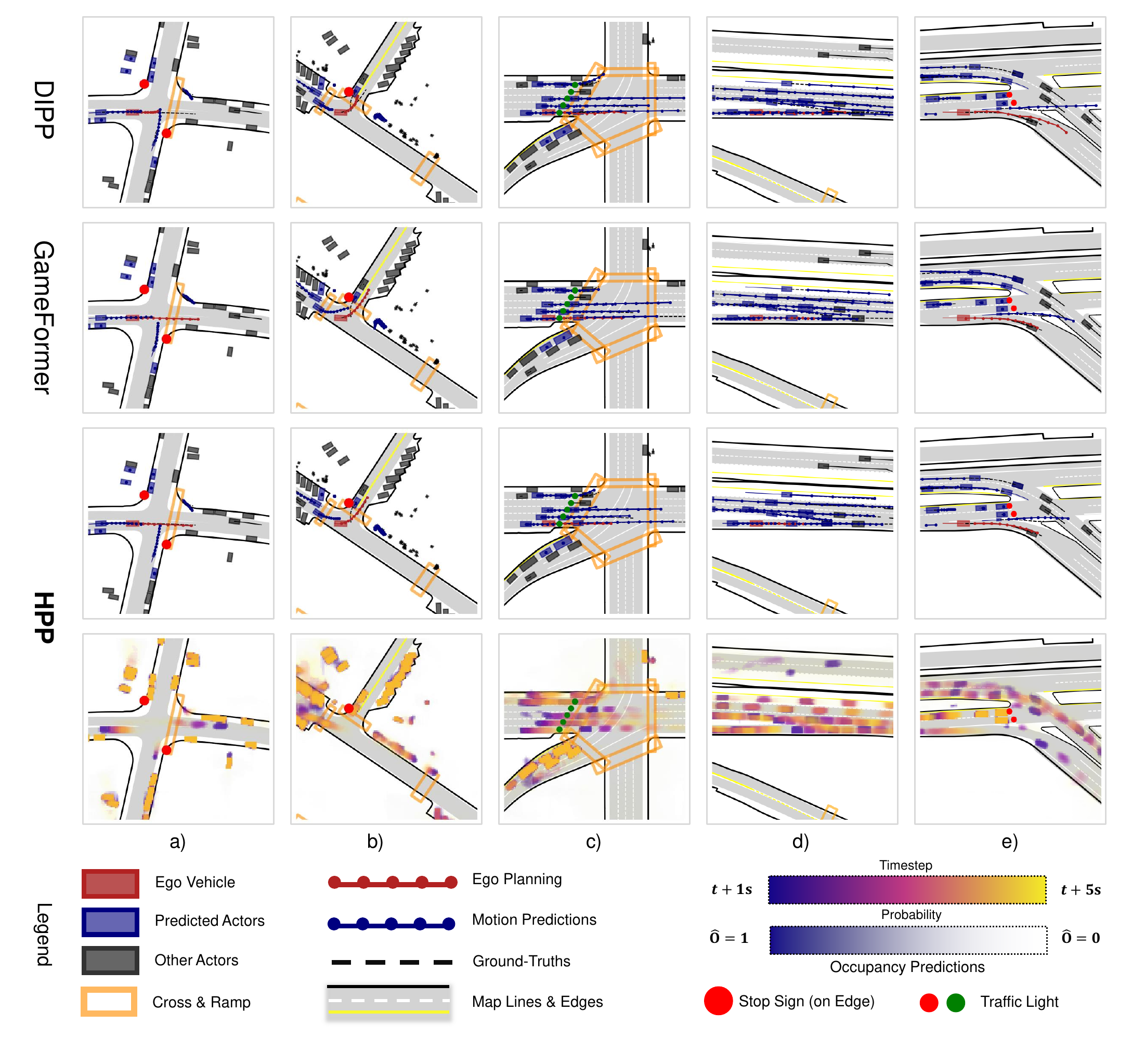}
    \caption{Qualitative comparison of HPP with SOTA integrated prediction and planning methods \cite{huang2022differentiable,huang2023gameformer} in WOMD \cite{ettinger2021large}. Performances integrated motion predictions (in blue), planning (red), and occupancy predictions (violet in separated sub-figures) are rendered with ground truth (black dotted lines) and scene context. All systems are compared with highly interactive scenarios: a) Crossing an unsignalized intersection with an incoming vehicle; b) Unprotected left-turn with a potential take-over vehicle; c) Cruising a five-point intersection with heavy traffic; d) Merging in a highway; and e) Right-turn with heading cyclist and low-speed vehicle.} 
    \label{fig8}
    \vspace{-0.5cm}
\end{figure*}

\subsection{Qualitative Results}
Figs. \ref{fig7} and \ref{fig8} showcase the qualitative benefits of integrating hybrid prediction in the nuScenes \cite{caesar2020nuscenes} and WOMD \cite{ettinger2021large} benchmarks, respectively. In the comprehensive testing scenario depicted in Fig. \ref{fig7}a, joint guidance from occupancy ensures consistent motion predictions during cruising. Notably, Fig. \ref{fig7}b demonstrates that hybrid prediction awareness contributes to consistent planning, enhancing reasoning behavior for smooth cruising without overly conservative avoidance, thereby mitigating risks near lane boundaries. Fig. \ref{fig7}c further illustrates the complementary effects, where motion predictions take precedence in guiding planning when occupancy is uncertain. The mutual influence is evident in Fig. \ref{fig7}d, where the ego vehicle remains in the lane instead of avoiding, as the occupancy predictions for rear cyclists are constrained by marginal predictions to follow a straight trajectory, maintaining a safe distance from the ego vehicle. 

In Fig. \ref{fig8}, qualitative assessments of WOMD interactive scenarios underscore HPP's superior reasoning capabilities and safety-conscious planning. In Fig. \ref{fig8}a, DIPP \cite{huang2022differentiable} encounters failure in an emergency stop, exposing a lack of consistency between motion predictions and planning, while HPP and GameFormer \cite{huang2023gameformer} exhibit smoother planning around obstacles. Motion prediction in HPP operates more smoothly with \cite{huang2023gameformer}, as occupancy predictions modulate the joint behaviors. This is important as the reasoned motion predictions \cite{huang2023gameformer} may trade consistency for safety (Fig. \ref{fig8}b) and accuracy (Fig. \ref{fig8}c). High-speed situations (Fig. \ref{fig8}d, Fig. \ref{fig8}e) further demonstrate HPP's mutual consistency between motion predictions and planning, showcasing its adaptability and robustness in diverse driving environments. These results underscore the efficacy of HPP in enhancing both the consistency and safety aspects of the autonomous driving system.

\subsection{Future Outlook}
In introducing HPP, we advocate for the adoption of modular co-design and optimization principles to shape ADS. As the notion of a planning-oriented modular learning system gains traction, it is crucial to underscore the significance of well-organized and concurrently integrated modules that mirror the complexities of real-world scenarios. In the future, we see agent-based models, such as LLMs, facilitating connections between various modules to ensure seamless integration. HPP lays the groundwork for module-wise integration, and future endeavors will delve deeper into exploring intermediate-level integration and guidance between prediction and planning. 

\section{Conclusions}
\label{sec5}
In this paper, we develop HPP, a modular co-design optimization framework for autonomous driving systems. The main focus of HPP is on integrating hybrid prediction and planning, wherein three sub-modules, i.e., MS-OccFormer, GTFormer, and Ego Planner address consistency issues and enhance adaptability using the hybrid prediction-guided learning and optimization pipeline. HPP has been extensively evaluated across diverse benchmarks, and the results consistently demonstrate its superior performance in both prediction and planning metrics compared to state-of-the-art methods.  The hybrid prediction awareness in HPP, which incorporates joint behaviors and occupancy predictions, improves qualitative consistency, safety, and feasibility in real-world scenarios. By elucidating the roles in modular co-design and the complementary effects that enhance planning through hybrid prediction, HPP showcases its potential to advance the field by tackling challenges in prediction and planning, contributing to the ongoing evolution of autonomous driving frameworks.

\begin{table}[tp]
\caption{Notations and parameters}
\centering
\resizebox{\linewidth}{!}{
\begin{tabular}{lllll}
\toprule
\multirow{2}{*}{Notation}& \multirow{2}{*}{Meaning}            & \multicolumn{3}{c}{Testing Benchmarks}  \\
                        &                                      & nuScenes & WOMD       & CARLA       \\ \midrule
$H,W$                   & Size of BEV space                    & 200     & 128/256    & 128         \\
                        & BEV length (m)                       & 50      & 64         & 64          \\
$N_A$                   & Max number of agents                 &  -      & 11/33      & 11          \\
$N_M$                   & Max number of map                    &  -      & 100        & 50          \\
$T$                     & Future Horizons (s)                  & 3/2/6   & 5/8        & 2           \\
                        & Future frequency (Hz)                & 2       & 2/1        & 2           \\
$M$                     & Number of modalities                 & 6       & 6          & 6           \\\midrule
$L$                     & Occ integration levels               & 3       & 3          & 3           \\
$K$                     & Reasoning levels                     & 3       & 3          & 3            \\
$L_P$                   & Layers in Ego Planner                & 3       & 3          & 3             \\
$h$                     & Number of attention heads            & 8       & 8          & 8            \\
$D$                     & Embedding size                       & 256     & 256        & 256          \\
                        & Activation function                  & \multicolumn{3}{c}{$\operatorname{ReLU}$}\\
                        & Dropout rate                         & 0.1     & 0.1        & 0.1\\\midrule
                        & Batch size                           & 4       & 24         & 32     \\
                        & Learning rate                        & $1e^{-4}$ &$1e^{-4}$    & $1e^{-4}$    \\
                        & Training epochs                      & 20      & 20        & 20 \\\bottomrule
\end{tabular}
}
\label{table12}
\end{table}

\bibliographystyle{IEEEtran}
\bibliography{b1}

% Generated by IEEEtran.bst, version: 1.14 (2015/08/26)
\begin{thebibliography}{10}
\providecommand{\url}[1]{#1}
\csname url@samestyle\endcsname
\providecommand{\newblock}{\relax}
\providecommand{\bibinfo}[2]{#2}
\providecommand{\BIBentrySTDinterwordspacing}{\spaceskip=0pt\relax}
\providecommand{\BIBentryALTinterwordstretchfactor}{4}
\providecommand{\BIBentryALTinterwordspacing}{\spaceskip=\fontdimen2\font plus
\BIBentryALTinterwordstretchfactor\fontdimen3\font minus
  \fontdimen4\font\relax}
\providecommand{\BIBforeignlanguage}[2]{{%
\expandafter\ifx\csname l@#1\endcsname\relax
\typeout{** WARNING: IEEEtran.bst: No hyphenation pattern has been}%
\typeout{** loaded for the language `#1'. Using the pattern for}%
\typeout{** the default language instead.}%
\else
\language=\csname l@#1\endcsname
\fi
#2}}
\providecommand{\BIBdecl}{\relax}
\BIBdecl

\bibitem{chen2022milestones}
L.~Chen, Y.~Li, C.~Huang, B.~Li, Y.~Xing, D.~Tian, L.~Li, Z.~Hu, X.~Na, Z.~Li
  \emph{et~al.}, ``Milestones in autonomous driving and intelligent vehicles:
  Survey of surveys,'' \emph{IEEE Transactions on Intelligent Vehicles}, 2022.

\bibitem{hu2023planning}
Y.~Hu, J.~Yang, L.~Chen, K.~Li, C.~Sima, X.~Zhu, S.~Chai, S.~Du, T.~Lin,
  W.~Wang \emph{et~al.}, ``Planning-oriented autonomous driving,'' in
  \emph{Proceedings of the IEEE/CVF Conference on Computer Vision and Pattern
  Recognition}, 2023, pp. 17\,853--17\,862.

\bibitem{huang2022differentiable}
Z.~Huang, H.~Liu, J.~Wu, and C.~Lv, ``Differentiable integrated motion
  prediction and planning with learnable cost function for autonomous
  driving,'' \emph{arXiv preprint arXiv:2207.10422}, 2022.

\bibitem{hu2022st}
S.~Hu, L.~Chen, P.~Wu, H.~Li, J.~Yan, and D.~Tao, ``St-p3: End-to-end
  vision-based autonomous driving via spatial-temporal feature learning,'' in
  \emph{European Conference on Computer Vision}.\hskip 1em plus 0.5em minus
  0.4em\relax Springer, 2022, pp. 533--549.

\bibitem{jiang2023vad}
B.~Jiang, S.~Chen, Q.~Xu, B.~Liao, J.~Chen, H.~Zhou, Q.~Zhang, W.~Liu,
  C.~Huang, and X.~Wang, ``Vad: Vectorized scene representation for efficient
  autonomous driving,'' \emph{arXiv preprint arXiv:2303.12077}, 2023.

\bibitem{hagedorn2023rethinking}
S.~Hagedorn, M.~Hallgarten, M.~Stoll, and A.~Condurache, ``Rethinking
  integration of prediction and planning in deep learning-based automated
  driving systems: a review,'' \emph{arXiv preprint arXiv:2308.05731}, 2023.

\bibitem{mo2022multi}
X.~Mo, Z.~Huang, Y.~Xing, and C.~Lv, ``Multi-agent trajectory prediction with
  heterogeneous edge-enhanced graph attention network,'' \emph{IEEE
  Transactions on Intelligent Transportation Systems}, 2022.

\bibitem{pini2023safe}
S.~Pini, C.~S. Perone, A.~Ahuja, A.~S.~R. Ferreira, M.~Niendorf, and
  S.~Zagoruyko, ``Safe real-world autonomous driving by learning to predict and
  plan with a mixture of experts,'' in \emph{2023 IEEE International Conference
  on Robotics and Automation (ICRA)}.\hskip 1em plus 0.5em minus 0.4em\relax
  IEEE, 2023, pp. 10\,069--10\,075.

\bibitem{liu2023occupancy}
H.~Liu, Z.~Huang, and C.~Lv, ``Occupancy prediction-guided neural planner for
  autonomous driving,'' \emph{arXiv preprint arXiv:2305.03303}, 2023.

\bibitem{hu2023imitation}
Y.~Hu, K.~Li, P.~Liang, J.~Qian, Z.~Yang, H.~Zhang, W.~Shao, Z.~Ding, W.~Xu,
  and Q.~Liu, ``Imitation with spatial-temporal heatmap: 2nd place solution for
  nuplan challenge,'' \emph{arXiv preprint arXiv:2306.15700}, 2023.

\bibitem{bansal2018chauffeurnet}
M.~Bansal, A.~Krizhevsky, and A.~Ogale, ``Chauffeurnet: Learning to drive by
  imitating the best and synthesizing the worst,'' \emph{arXiv preprint
  arXiv:1812.03079}, 2018.

\bibitem{mozaffari2020deep}
S.~Mozaffari, O.~Y. Al-Jarrah, M.~Dianati, P.~Jennings, and A.~Mouzakitis,
  ``Deep learning-based vehicle behavior prediction for autonomous driving
  applications: A review,'' \emph{IEEE Transactions on Intelligent
  Transportation Systems}, vol.~23, no.~1, pp. 33--47, 2020.

\bibitem{kim2022stopnet}
J.~Kim, R.~Mahjourian, S.~Ettinger, M.~Bansal, B.~White, B.~Sapp, and
  D.~Anguelov, ``Stopnet: Scalable trajectory and occupancy prediction for
  urban autonomous driving,'' \emph{arXiv preprint arXiv:2206.00991}, 2022.

\bibitem{chen2023end}
L.~Chen, P.~Wu, K.~Chitta, B.~Jaeger, A.~Geiger, and H.~Li, ``End-to-end
  autonomous driving: Challenges and frontiers,'' \emph{arXiv preprint
  arXiv:2306.16927}, 2023.

\bibitem{ye2023fusionad}
T.~Ye, W.~Jing, C.~Hu, S.~Huang, L.~Gao, F.~Li, J.~Wang, K.~Guo, W.~Xiao,
  W.~Mao \emph{et~al.}, ``Fusionad: Multi-modality fusion for prediction and
  planning tasks of autonomous driving,'' \emph{arXiv preprint
  arXiv:2308.01006}, 2023.

\bibitem{jia2023hdgt}
X.~Jia, P.~Wu, L.~Chen, Y.~Liu, H.~Li, and J.~Yan, ``Hdgt: Heterogeneous
  driving graph transformer for multi-agent trajectory prediction via scene
  encoding,'' \emph{IEEE transactions on pattern analysis and machine
  intelligence}, 2023.

\bibitem{huang2022multi}
Z.~Huang, X.~Mo, and C.~Lv, ``Multi-modal motion prediction with
  transformer-based neural network for autonomous driving,'' in \emph{2022
  International Conference on Robotics and Automation (ICRA)}.\hskip 1em plus
  0.5em minus 0.4em\relax IEEE, 2022, pp. 2605--2611.

\bibitem{mo2023map}
X.~Mo, Y.~Xing, H.~Liu, and C.~Lv, ``Map-adaptive multimodal trajectory
  prediction using hierarchical graph neural networks,'' \emph{IEEE Robotics
  and Automation Letters}, 2023.

\bibitem{shi2022motion}
S.~Shi, L.~Jiang, D.~Dai, and B.~Schiele, ``Motion transformer with global
  intention localization and local movement refinement,'' \emph{Advances in
  Neural Information Processing Systems}, 2022.

\bibitem{hu2022hope}
Y.~Hu, W.~Shao, B.~Jiang, J.~Chen, S.~Chai, Z.~Yang, J.~Qian, H.~Zhou, and
  Q.~Liu, ``Hope: Hierarchical spatial-temporal network for occupancy flow
  prediction,'' \emph{arXiv preprint arXiv:2206.10118}, 2022.

\bibitem{huang2022vectorflow}
X.~Huang, X.~Tian, J.~Gu, Q.~Sun, and H.~Zhao, ``Vectorflow: Combining images
  and vectors for traffic occupancy and flow prediction,'' \emph{arXiv preprint
  arXiv:2208.04530}, 2022.

\bibitem{gilles2021thomas}
T.~Gilles, S.~Sabatini, D.~Tsishkou, B.~Stanciulescu, and F.~Moutarde,
  ``Thomas: Trajectory heatmap output with learned multi-agent sampling,''
  \emph{arXiv preprint arXiv:2110.06607}, 2021.

\bibitem{kamenev2022predictionnet}
A.~Kamenev, L.~Wang, O.~B. Bohan, I.~Kulkarni, B.~Kartal, A.~Molchanov,
  S.~Birchfield, D.~Nist{\'e}r, and N.~Smolyanskiy, ``Predictionnet: Real-time
  joint probabilistic traffic prediction for planning, control, and
  simulation,'' in \emph{2022 International Conference on Robotics and
  Automation (ICRA)}.\hskip 1em plus 0.5em minus 0.4em\relax IEEE, 2022, pp.
  8936--8942.

\bibitem{huang2023conditional}
Z.~Huang, H.~Liu, J.~Wu, and C.~Lv, ``Conditional predictive behavior planning
  with inverse reinforcement learning for human-like autonomous driving,''
  \emph{IEEE Transactions on Intelligent Transportation Systems}, 2023.

\bibitem{hang2020integrated}
P.~Hang, C.~Lv, C.~Huang, J.~Cai, Z.~Hu, and Y.~Xing, ``An integrated framework
  of decision making and motion planning for autonomous vehicles considering
  social behaviors,'' \emph{IEEE transactions on vehicular technology},
  vol.~69, no.~12, pp. 14\,458--14\,469, 2020.

\bibitem{xu2023bits}
D.~Xu, Y.~Chen, B.~Ivanovic, and M.~Pavone, ``Bits: Bi-level imitation for
  traffic simulation,'' in \emph{2023 IEEE International Conference on Robotics
  and Automation (ICRA)}.\hskip 1em plus 0.5em minus 0.4em\relax IEEE, 2023,
  pp. 2929--2936.

\bibitem{liu2022augmenting}
H.~Liu, Z.~Huang, X.~Mo, and C.~Lv, ``Augmenting reinforcement learning with
  transformer-based scene representation learning for decision-making of
  autonomous driving,'' \emph{arXiv preprint arXiv:2208.12263}, 2022.

\bibitem{renz2022plant}
K.~Renz, K.~Chitta, O.-B. Mercea, A.~Koepke, Z.~Akata, and A.~Geiger, ``Plant:
  Explainable planning transformers via object-level representations,''
  \emph{arXiv preprint arXiv:2210.14222}, 2022.

\bibitem{rhinehart2019precog}
N.~Rhinehart, R.~McAllister, K.~Kitani, and S.~Levine, ``Precog: Prediction
  conditioned on goals in visual multi-agent settings,'' in \emph{Proceedings
  of the IEEE/CVF International Conference on Computer Vision}, 2019, pp.
  2821--2830.

\bibitem{huang2023learning}
Z.~Huang, H.~Liu, J.~Wu, W.~Huang, and C.~Lv, ``Learning interaction-aware
  motion prediction model for decision-making in autonomous driving,''
  \emph{arXiv preprint arXiv:2302.03939}, 2023.

\bibitem{espinoza2022deep}
J.~L.~V. Espinoza, A.~Liniger, W.~Schwarting, D.~Rus, and L.~Van~Gool, ``Deep
  interactive motion prediction and planning: Playing games with motion
  prediction models,'' in \emph{Learning for Dynamics and Control
  Conference}.\hskip 1em plus 0.5em minus 0.4em\relax PMLR, 2022, pp.
  1006--1019.

\bibitem{burger2022interaction}
C.~Burger, J.~Fischer, F.~Bieder, {\"O}.~{\c{S}}. Ta{\c{s}}, and C.~Stiller,
  ``Interaction-aware game-theoretic motion planning for automated vehicles
  using bi-level optimization,'' in \emph{2022 IEEE 25th International
  Conference on Intelligent Transportation Systems (ITSC)}.\hskip 1em plus
  0.5em minus 0.4em\relax IEEE, 2022, pp. 3978--3985.

\bibitem{wang2022social}
W.~Wang, L.~Wang, C.~Zhang, C.~Liu, L.~Sun \emph{et~al.}, ``Social interactions
  for autonomous driving: A review and perspectives,'' \emph{Foundations and
  Trends{\textregistered} in Robotics}, vol.~10, no. 3-4, pp. 198--376, 2022.

\bibitem{huang2023gameformer}
Z.~Huang, H.~Liu, and C.~Lv, ``Gameformer: Game-theoretic modeling and learning
  of transformer-based interactive prediction and planning for autonomous
  driving,'' \emph{arXiv preprint arXiv:2303.05760}, 2023.

\bibitem{karkus2023diffstack}
P.~Karkus, B.~Ivanovic, S.~Mannor, and M.~Pavone, ``Diffstack: A differentiable
  and modular control stack for autonomous vehicles,'' in \emph{Conference on
  Robot Learning}.\hskip 1em plus 0.5em minus 0.4em\relax PMLR, 2023, pp.
  2170--2180.

\bibitem{hanselmann2022king}
N.~Hanselmann, K.~Renz, K.~Chitta, A.~Bhattacharyya, and A.~Geiger, ``King:
  Generating safety-critical driving scenarios for robust imitation via
  kinematics gradients,'' in \emph{European Conference on Computer
  Vision}.\hskip 1em plus 0.5em minus 0.4em\relax Springer, 2022, pp. 335--352.

\bibitem{casas2021mp3}
S.~Casas, A.~Sadat, and R.~Urtasun, ``Mp3: A unified model to map, perceive,
  predict and plan,'' in \emph{Proceedings of the IEEE/CVF Conference on
  Computer Vision and Pattern Recognition}, 2021, pp. 14\,403--14\,412.

\bibitem{liang2020pnpnet}
M.~Liang, B.~Yang, W.~Zeng, Y.~Chen, R.~Hu, S.~Casas, and R.~Urtasun, ``Pnpnet:
  End-to-end perception and prediction with tracking in the loop,'' in
  \emph{Proceedings of the IEEE/CVF Conference on Computer Vision and Pattern
  Recognition}, 2020, pp. 11\,553--11\,562.

\bibitem{li2022bevformer}
Z.~Li, W.~Wang, H.~Li, E.~Xie, C.~Sima, T.~Lu, Y.~Qiao, and J.~Dai,
  ``Bevformer: Learning bird’s-eye-view representation from multi-camera
  images via spatiotemporal transformers,'' in \emph{European conference on
  computer vision}.\hskip 1em plus 0.5em minus 0.4em\relax Springer, 2022, pp.
  1--18.

\bibitem{zhang2022beverse}
Y.~Zhang, Z.~Zhu, W.~Zheng, J.~Huang, G.~Huang, J.~Zhou, and J.~Lu, ``Beverse:
  Unified perception and prediction in birds-eye-view for vision-centric
  autonomous driving,'' \emph{arXiv preprint arXiv:2205.09743}, 2022.

\bibitem{akan2022stretchbev}
A.~K. Akan and F.~G{\"u}ney, ``Stretchbev: Stretching future instance
  prediction spatially and temporally,'' in \emph{European Conference on
  Computer Vision}.\hskip 1em plus 0.5em minus 0.4em\relax Springer, 2022, pp.
  444--460.

\bibitem{li2023delving}
H.~Li, C.~Sima, J.~Dai, W.~Wang, L.~Lu, H.~Wang, J.~Zeng, Z.~Li, J.~Yang,
  H.~Deng \emph{et~al.}, ``Delving into the devils of bird's-eye-view
  perception: A review, evaluation and recipe,'' \emph{IEEE Transactions on
  Pattern Analysis and Machine Intelligence}, 2023.

\bibitem{jia2023driveadapter}
X.~Jia, Y.~Gao, L.~Chen, J.~Yan, P.~L. Liu, and H.~Li, ``Driveadapter: Breaking
  the coupling barrier of perception and planning in end-to-end autonomous
  driving,'' in \emph{Proceedings of the IEEE/CVF International Conference on
  Computer Vision}, 2023, pp. 7953--7963.

\bibitem{mao2023gpt}
J.~Mao, Y.~Qian, H.~Zhao, and Y.~Wang, ``Gpt-driver: Learning to drive with
  gpt,'' \emph{arXiv preprint arXiv:2310.01415}, 2023.

\bibitem{mao2023language}
J.~Mao, J.~Ye, Y.~Qian, M.~Pavone, and Y.~Wang, ``A language agent for
  autonomous driving,'' \emph{arXiv preprint arXiv:2311.10813}, 2023.

\bibitem{sima2023drivelm}
C.~Sima, K.~Renz, K.~Chitta, L.~Chen, H.~Zhang, C.~Xie, P.~Luo, A.~Geiger, and
  H.~Li, ``Drivelm: Driving with graph visual question answering,'' \emph{arXiv
  preprint arXiv:2312.14150}, 2023.

\bibitem{he2016deep}
K.~He, X.~Zhang, S.~Ren, and J.~Sun, ``Deep residual learning for image
  recognition,'' in \emph{Proceedings of the IEEE conference on computer vision
  and pattern recognition}, 2016, pp. 770--778.

\bibitem{liu2023multi}
H.~Liu, Z.~Huang, and C.~Lv, ``Multi-modal hierarchical transformer for
  occupancy flow field prediction in autonomous driving,'' in \emph{2023 IEEE
  International Conference on Robotics and Automation (ICRA)}.\hskip 1em plus
  0.5em minus 0.4em\relax IEEE, 2023, pp. 1449--1455.

\bibitem{liu2021swin}
Z.~Liu, Y.~Lin, Y.~Cao, H.~Hu, Y.~Wei, Z.~Zhang, S.~Lin, and B.~Guo, ``Swin
  transformer: Hierarchical vision transformer using shifted windows,'' in
  \emph{Proceedings of the IEEE/CVF International Conference on Computer
  Vision}, 2021, pp. 10\,012--10\,022.

\bibitem{bhardwaj2020differentiable}
M.~Bhardwaj, B.~Boots, and M.~Mukadam, ``Differentiable gaussian process motion
  planning,'' in \emph{2020 IEEE international conference on robotics and
  automation (ICRA)}.\hskip 1em plus 0.5em minus 0.4em\relax IEEE, 2020, pp.
  10\,598--10\,604.

\bibitem{caesar2020nuscenes}
H.~Caesar, V.~Bankiti, A.~H. Lang, S.~Vora, V.~E. Liong, Q.~Xu, A.~Krishnan,
  Y.~Pan, G.~Baldan, and O.~Beijbom, ``nuscenes: A multimodal dataset for
  autonomous driving,'' in \emph{Proceedings of the IEEE/CVF conference on
  computer vision and pattern recognition}, 2020, pp. 11\,621--11\,631.

\bibitem{ettinger2021large}
S.~Ettinger, S.~Cheng, B.~Caine, C.~Liu, H.~Zhao, S.~Pradhan, Y.~Chai, B.~Sapp,
  C.~R. Qi, Y.~Zhou \emph{et~al.}, ``Large scale interactive motion forecasting
  for autonomous driving: The waymo open motion dataset,'' in \emph{Proceedings
  of the IEEE/CVF International Conference on Computer Vision}, 2021, pp.
  9710--9719.

\bibitem{mahjourian2022occupancy}
R.~Mahjourian, J.~Kim, Y.~Chai, M.~Tan, B.~Sapp, and D.~Anguelov, ``Occupancy
  flow fields for motion forecasting in autonomous driving,'' \emph{IEEE
  Robotics and Automation Letters}, vol.~7, no.~2, pp. 5639--5646, 2022.

\bibitem{zeng2019end}
W.~Zeng, W.~Luo, S.~Suo, A.~Sadat, B.~Yang, S.~Casas, and R.~Urtasun,
  ``End-to-end interpretable neural motion planner,'' in \emph{Proceedings of
  the IEEE/CVF Conference on Computer Vision and Pattern Recognition}, 2019,
  pp. 8660--8669.

\bibitem{hu2021safe}
P.~Hu, A.~Huang, J.~Dolan, D.~Held, and D.~Ramanan, ``Safe local motion
  planning with self-supervised freespace forecasting,'' in \emph{Proceedings
  of the IEEE/CVF Conference on Computer Vision and Pattern Recognition}, 2021,
  pp. 12\,732--12\,741.

\bibitem{khurana2022differentiable}
T.~Khurana, P.~Hu, A.~Dave, J.~Ziglar, D.~Held, and D.~Ramanan,
  ``Differentiable raycasting for self-supervised occupancy forecasting,'' in
  \emph{European Conference on Computer Vision}.\hskip 1em plus 0.5em minus
  0.4em\relax Springer, 2022, pp. 353--369.

\bibitem{tong2023scene}
W.~Tong, C.~Sima, T.~Wang, L.~Chen, S.~Wu, H.~Deng, Y.~Gu, L.~Lu, P.~Luo,
  D.~Lin \emph{et~al.}, ``Scene as occupancy,'' in \emph{Proceedings of the
  IEEE/CVF International Conference on Computer Vision}, 2023, pp. 8406--8415.

\bibitem{chen2023deepemplanner}
Z.~Chen, M.~Ye, S.~Xu, T.~Cao, and Q.~Chen, ``Deepemplanner: An em motion
  planner with iterative interactions,'' \emph{arXiv preprint
  arXiv:2311.08100}, 2023.

\bibitem{dosovitskiy2017carla}
A.~Dosovitskiy, G.~Ros, F.~Codevilla, A.~Lopez, and V.~Koltun, ``Carla: An open
  urban driving simulator,'' in \emph{Conference on robot learning}.\hskip 1em
  plus 0.5em minus 0.4em\relax PMLR, 2017, pp. 1--16.

\bibitem{chitta2022transfuser}
K.~Chitta, A.~Prakash, B.~Jaeger, Z.~Yu, K.~Renz, and A.~Geiger, ``Transfuser:
  Imitation with transformer-based sensor fusion for autonomous driving,''
  \emph{IEEE Transactions on Pattern Analysis and Machine Intelligence}, 2022.

\bibitem{chen2017deeplab}
L.-C. Chen, G.~Papandreou, I.~Kokkinos, K.~Murphy, and A.~L. Yuille, ``Deeplab:
  Semantic image segmentation with deep convolutional nets, atrous convolution,
  and fully connected crfs,'' \emph{IEEE transactions on pattern analysis and
  machine intelligence}, vol.~40, no.~4, pp. 834--848, 2017.

\bibitem{kim2020video}
D.~Kim, S.~Woo, J.-Y. Lee, and I.~S. Kweon, ``Video panoptic segmentation,'' in
  \emph{Proceedings of the IEEE/CVF Conference on Computer Vision and Pattern
  Recognition}, 2020, pp. 9859--9868.

\bibitem{gu2023vip3d}
J.~Gu, C.~Hu, T.~Zhang, X.~Chen, Y.~Wang, Y.~Wang, and H.~Zhao, ``Vip3d:
  End-to-end visual trajectory prediction via 3d agent queries,'' in
  \emph{Proceedings of the IEEE/CVF Conference on Computer Vision and Pattern
  Recognition}, 2023, pp. 5496--5506.

\bibitem{scheel2022urban}
O.~Scheel, L.~Bergamini, M.~Wolczyk, B.~Osi{\'n}ski, and P.~Ondruska, ``Urban
  driver: Learning to drive from real-world demonstrations using policy
  gradients,'' in \emph{Conference on Robot Learning}.\hskip 1em plus 0.5em
  minus 0.4em\relax PMLR, 2022, pp. 718--728.

\bibitem{liu2021multimodal}
Y.~Liu, J.~Zhang, L.~Fang, Q.~Jiang, and B.~Zhou, ``Multimodal motion
  prediction with stacked transformers,'' in \emph{Proceedings of the IEEE/CVF
  Conference on Computer Vision and Pattern Recognition}, 2021, pp. 7577--7586.

\bibitem{hu2021fiery}
A.~Hu, Z.~Murez, N.~Mohan, S.~Dudas, J.~Hawke, V.~Badrinarayanan, R.~Cipolla,
  and A.~Kendall, ``Fiery: Future instance prediction in bird's-eye view from
  surround monocular cameras,'' in \emph{Proceedings of the IEEE/CVF
  International Conference on Computer Vision}, 2021, pp. 15\,273--15\,282.

\bibitem{li2023powerbev}
P.~Li, S.~Ding, X.~Chen, N.~Hanselmann, M.~Cordts, and J.~Gall, ``Powerbev: A
  powerful yet lightweight framework for instance prediction in bird's-eye
  view,'' \emph{arXiv preprint arXiv:2306.10761}, 2023.

\bibitem{rhinehart2018deep}
N.~Rhinehart, R.~McAllister, and S.~Levine, ``Deep imitative models for
  flexible inference, planning, and control,'' \emph{arXiv preprint
  arXiv:1810.06544}, 2018.

\bibitem{varadarajan2022multipath++}
B.~Varadarajan, A.~Hefny, A.~Srivastava, K.~S. Refaat, N.~Nayakanti,
  A.~Cornman, K.~Chen, B.~Douillard, C.~P. Lam, D.~Anguelov \emph{et~al.},
  ``Multipath++: Efficient information fusion and trajectory aggregation for
  behavior prediction,'' in \emph{2022 International Conference on Robotics and
  Automation (ICRA)}.\hskip 1em plus 0.5em minus 0.4em\relax IEEE, 2022, pp.
  7814--7821.

\bibitem{kumar2020conservative}
A.~Kumar, A.~Zhou, G.~Tucker, and S.~Levine, ``Conservative q-learning for
  offline reinforcement learning,'' \emph{Advances in Neural Information
  Processing Systems}, vol.~33, pp. 1179--1191, 2020.

\bibitem{zhang2021end}
Z.~Zhang, A.~Liniger, D.~Dai, F.~Yu, and L.~Van~Gool, ``End-to-end urban
  driving by imitating a reinforcement learning coach,'' in \emph{Proceedings
  of the IEEE/CVF international conference on computer vision}, 2021, pp.
  15\,222--15\,232.

\end{thebibliography}

\end{document}